\documentclass[11pt,a4paper]{article}

\usepackage{graphicx}

\usepackage{subfigure}
\usepackage{setspace}
\usepackage[usenames,dvipsnames]{xcolor}
\usepackage{amssymb}
\usepackage{lineno}
\usepackage{tikz}
\usepackage{minibox}
\usepackage[top=2.5cm, bottom=2.5cm, left=2cm, right=2cm]{geometry}
\usepackage[colorlinks=false,pagebackref=false,urlcolor=Emerald,citecolor=blue,linkcolor=red,hypertexnames=false]{hyperref}
\usepackage{xr}
\usepackage{rotating}
\usepackage{soul}

\newcommand{\cc}[1][]{}
\newcommand{\chk}[1][]{}

\newcommand{\refstab}[1]{\textbf{Table \ref{tab-#1}}}
\newcommand{\refig}[1]{\textbf{Figure \ref{fig-#1}}}
\newcommand{\refsfig}[1]{\textbf{Figure \ref{fig-#1}}}
\newcommand{\refm}{\textbf{Methods}}

\newcommand{\cm}[1]{}

\newcommand*\circled[1]{\tikz[baseline=(char.base)]{
            \node[fill=red,shape=circle,inner sep=0pt,minimum size=1.4em] (char) {#1};}}

\graphicspath{{figures/}{figs/}{./}{../figures/}}
\def\titleorig{Wisdom of the crowd from unsupervised dimension reduction}
\def\authors{{\large
  Lingfei Wang\footnote{Corresponding author. Email: Lingfei.Wang@roslin.ed.ac.uk} and Tom Michoel

  Division of Genetics and Genomics, The Roslin Institute, The University of Edinburgh, Easter Bush, Midlothian EH25 9RG, UK \\
}}

\begin{document}

\begin{spacing}{1}\LARGE \textbf{\titleorig}\\[-8mm]\end{spacing}

\authors
\section*{Abstract}
Wisdom of the crowd, the collective intelligence derived from responses of multiple human or machine individuals to the same questions, can be more accurate than each individual, and improve social decision-making and prediction accuracy (\cite{galton_vox_1907,romney_culture_1986,marbach_wisdom_2012,parisi_ranking_2014,alhamdoosh_combining_2017}). This can also integrate multiple programs or datasets, each as an individual, for the same predictive questions. Crowd wisdom estimates each individual's independent error level arising from their limited knowledge, and finds the crowd consensus that minimizes the overall error. However, previous studies have merely built isolated, problem-specific models with limited generalizability, and mainly for binary (yes/no) responses. Here we show with simulation and real-world data that the crowd wisdom problem is analogous to one-dimensional unsupervised dimension reduction in machine learning. This provides a natural class of crowd wisdom solutions, such as principal component analysis and Isomap, which can handle binary and also continuous responses, like confidence levels, and consequently can be more accurate than existing solutions. They can even outperform supervised-learning-based collective intelligence that is calibrated on historical performance of individuals, \textit{e.g.}\ penalized linear regression and random forest. This study unifies crowd wisdom and unsupervised dimension reduction, and thereupon introduces a broad range of highly-performing and widely-applicable crowd wisdom methods. As the costs for data acquisition and processing rapidly decrease\cc, this study will promote and guide crowd wisdom applications in the social and natural sciences, including data fusion (\cite{cho_compact_2016}), meta-analysis (\cite{alhamdoosh_combining_2017}), crowd-sourcing (\cite{marbach_wisdom_2012,sheshadri_square:_2013}), and committee decision making (\cite{romney_culture_1986,chen_eliminating_2004}).

\section*{Results}
\begin{figure}
\begin{center}
\begin{tabular}{p{0em}p{0.38\linewidth}p{0em}p{0.38\linewidth}}
\vspace{0pt}\textbf{{\large A}}&\vspace{0pt}\includegraphics[width=\linewidth]{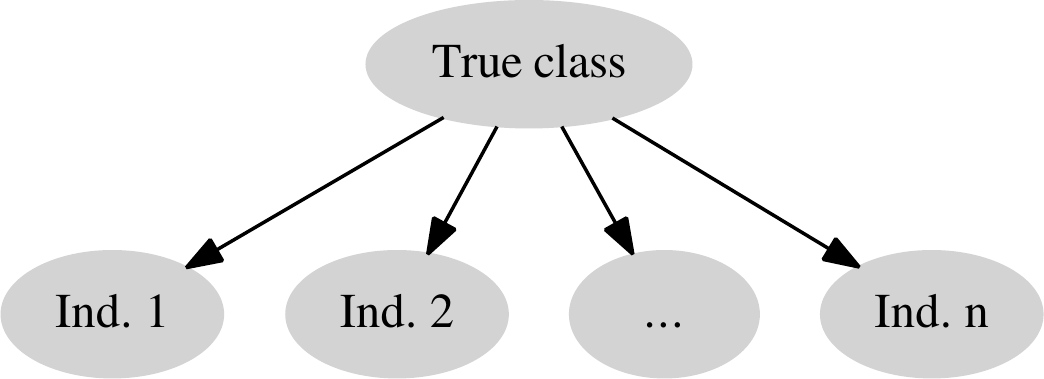}
&\vspace{0pt}\textbf{{\large B}}&\vspace{0pt}\includegraphics[width=\linewidth]{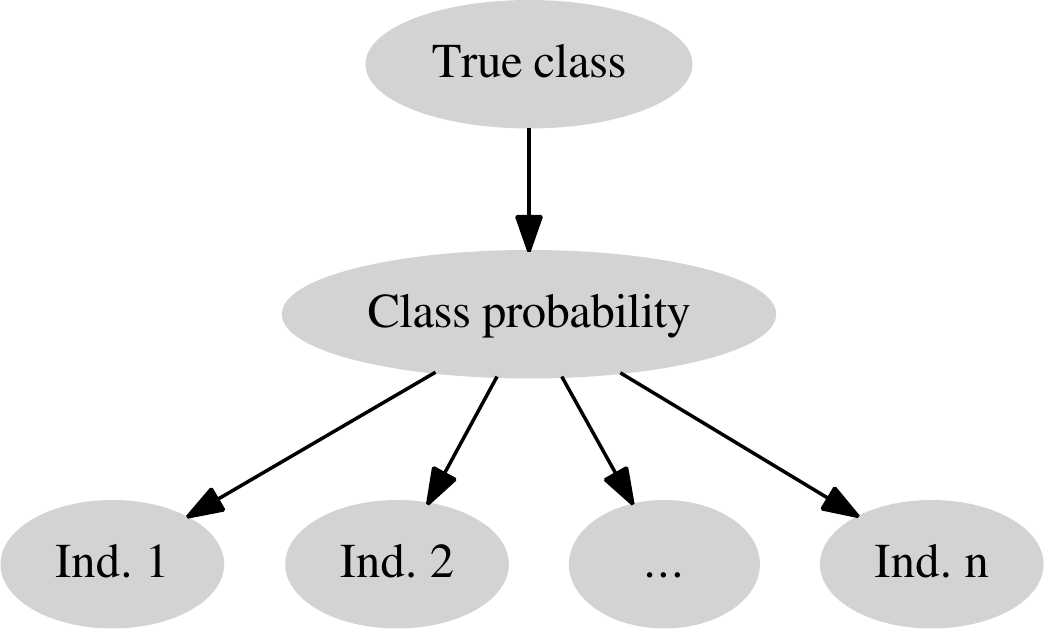}
\end{tabular}
\\
\begin{tabular}{p{0em}p{0.24\linewidth}p{0em}p{0.25\linewidth}p{0em}p{0.25\linewidth}}
\vspace{0pt}\textbf{{\large C}}&\vspace{0pt}\includegraphics[width=\linewidth]{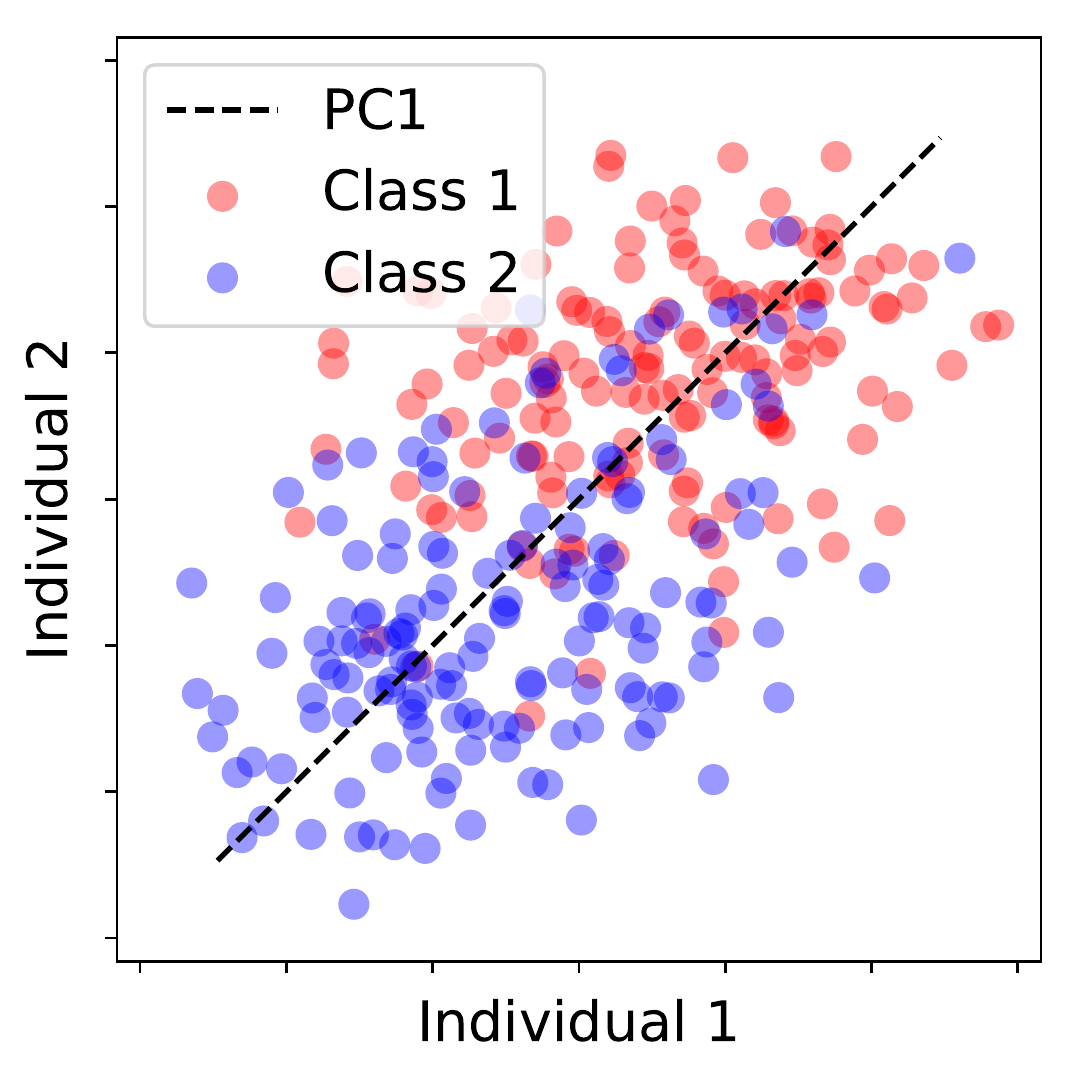}
&\vspace{0pt}\textbf{{\large D}}&\vspace{0pt}\includegraphics[width=\linewidth]{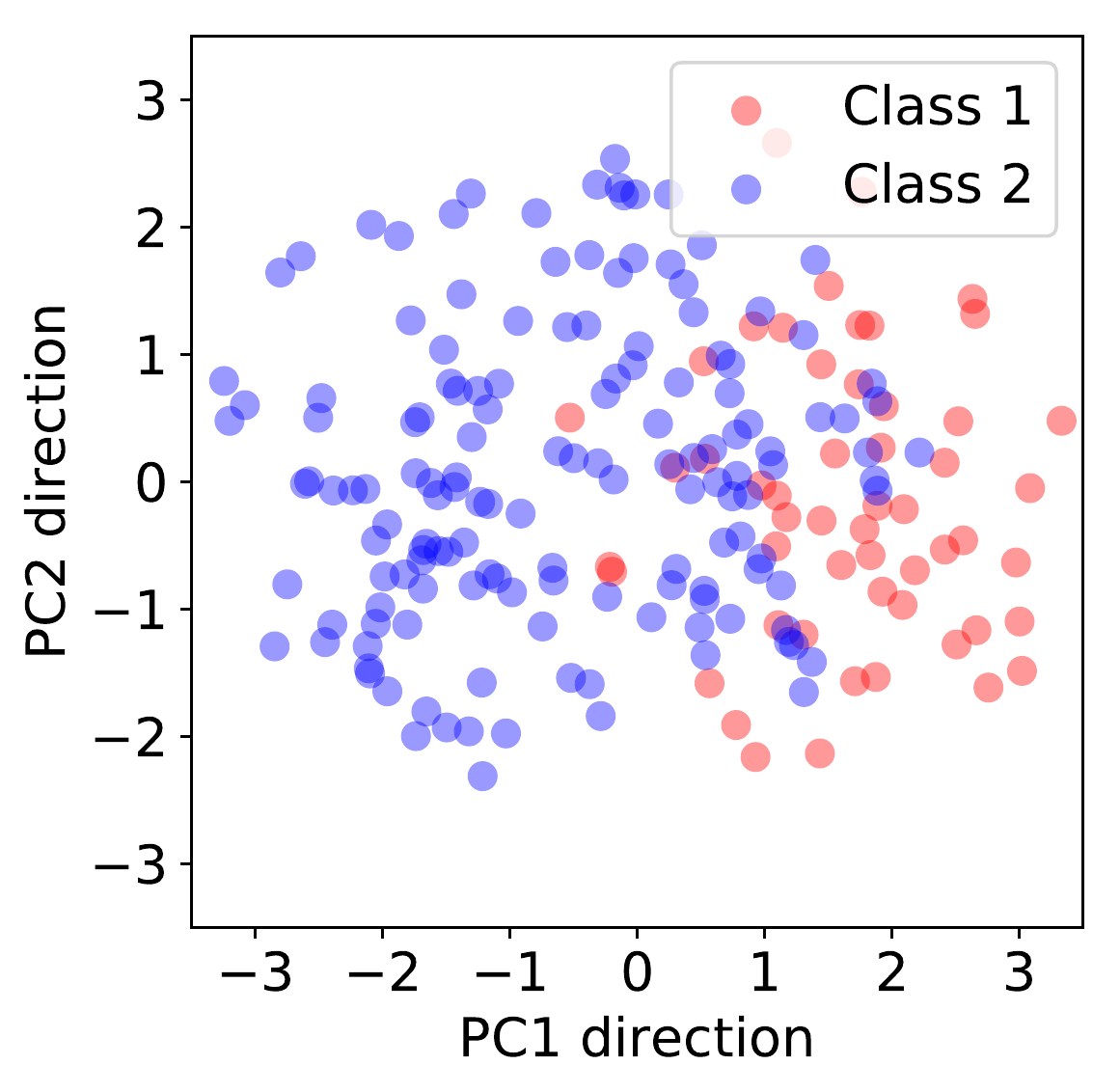}
&\vspace{0pt}\textbf{{\large E}}&\vspace{0pt}\includegraphics[width=\linewidth]{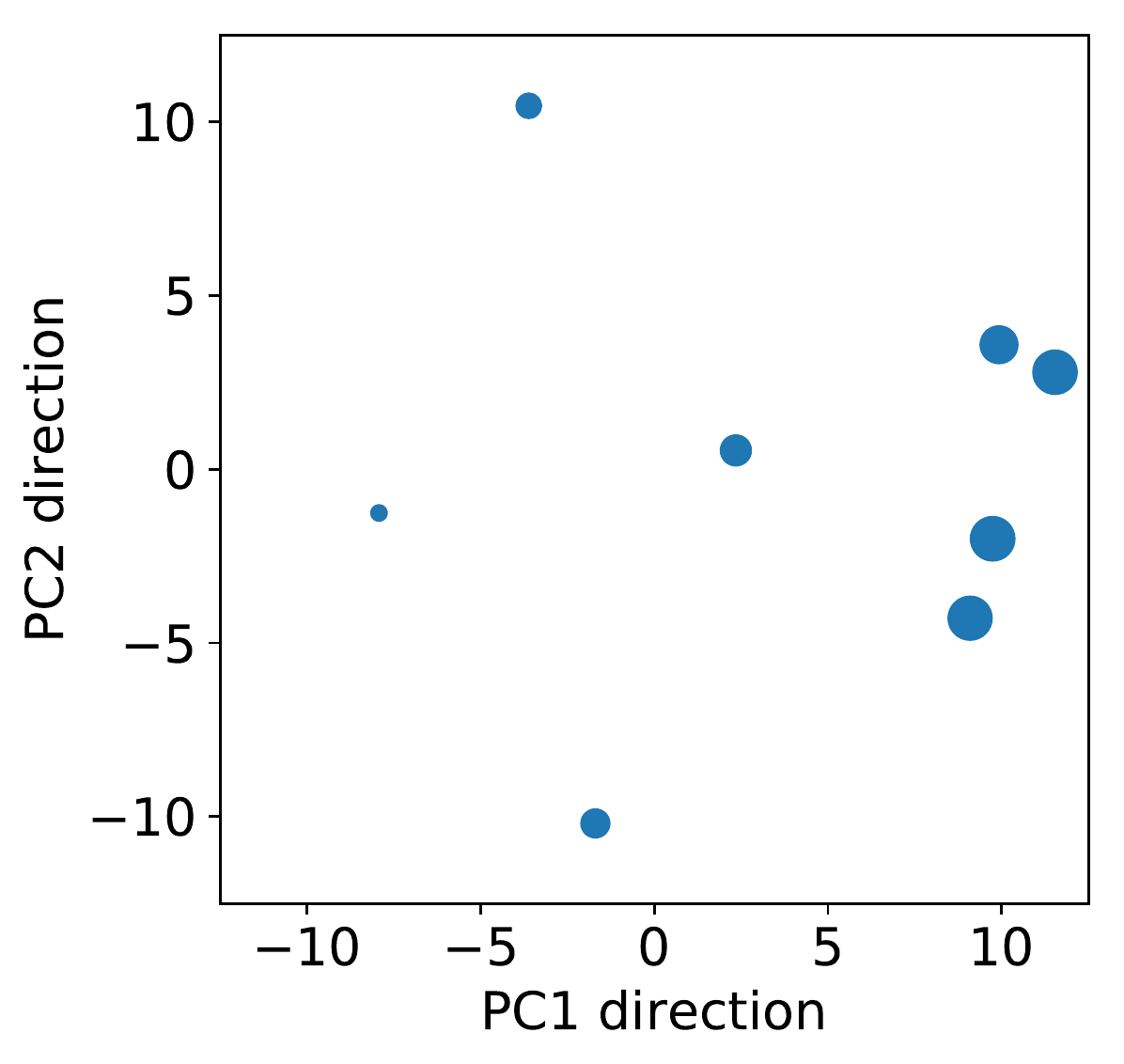}
\end{tabular}
\end{center}
\caption{\textbf{Illustrations of wisdom of the crowd.} \textbf{(A)} Probablistic graph of the conventional crowd wisdom. Each individual is assumed to contain independent errors on top of the true class. \textbf{(B)} Probablistic graph of the new crowd wisdom. The intermediate continuous variable of class probability is introduced as what individuals estimate with independent errors. \textbf{(C)} Illustrative application of PCA crowd wisdom on two individuals independently estimating the class probability. \textbf{(D, E)} PCA recovered classification (\textbf{D}) and individual accuracy (\textbf{E}, in terms of AUROC as radius) in PC1 direction the on DREAM2 dataset.\label{fig-1}}
\end{figure}

\chk[Update figure coloring]
\chk[SML converge to thresholded PCA in theory as individual count go to infinity]
Although wisdom of the crowd and its philosophy have been discovered and rediscovered in a wide range of sociological and statistical contexts, most studies rely on the fundamental assumption that each individual is an independent estimator of the ground-truth, possessing their knowledge as the signal and bias as the error (\refig{1}\textbf{A}). As long as the group or ensemble of individuals remain unbiased as a whole, aggregating individual estimators for the same predictive variables would still strengthen the signal and cancel out their errors. This can be regarded as a more complex version of averaging multiple measurements of the same variable.

However, previous crowd wisdom classification studies have focused predominantly on binary responses and problem-specific models (\cite{dawid_maximum_1979, sheshadri_square:_2013,parisi_ranking_2014,jaffe_estimating_2015}). The confusion matrices of individuals and the binary true classes are fit in turn to maximize the model's likelihood with expectation-maximization. Where available, continuous individual predictions such as confidence levels are thresholded and mostly lost, potentially limiting the classification accuracy and generalizability, whilst the proper choice of threshold can also be difficult.

To resolve this issue and to link crowd wisdom with machine learning, we consider continuous rather than binary variables for individual responses. Due to a lack of complete information to perfectly determine the true class, we introduce an unknown intermediate layer representing the probability of the true class (\textit{class probability}, \refig{1}\textbf{B}). In the simplest scenario, individual responses are then independent continuous estimations of the class probability. More generally, individuals can also characterize and estimate classification confidence with any other continuous scores, which are assumed to be equivalent in ranking with the class probability. Binary responses can also be treated as numerical 0s and 1s.

The continuous crowd wisdom classification problem can then be solved by unsupervised dimension reduction. Unsupervised dimension reduction infers the latent lower dimensions by which the input data are assumed to be parameterized. In crowd wisdom (\refig{1}\textbf{B}), each individual independently estimates, and is effectively parameterized by, the class probability alone. Therefore, the class probability may be recovered as the first and only dimension (\refig{1}\textbf{C}, subjecting to a monotonic transformation). This makes dimension reduction the natural crowd wisdom for classification problems with continuous information. Which dimension reduction method is the best then depends on various aspects of the problem, such as nonlinearity. As a brief demonstration with the DREAM2 BCL6 Transcription Factor Prediction challenge dataset, containing the confidence scores of 200 genes as potential targets of BCL6 (i.e.\ questions) submitted by 8 teams (i.e.\ individuals) (\cite{comm_gustavo_vogel_20167,noauthor_dream2_nodate,klein_transcriptional_2003}), the first principal component (PC1) direction of the gene-by-individual matrix gave an accurate representation of the class probability ranking (\refig{1}\textbf{D}) and the performance of each individual (\refig{1}\textbf{E}).

\begin{figure}
\begin{center}
\begin{tabular}{p{0em}p{0.32\linewidth}p{0em}p{0.32\linewidth}p{0.2\linewidth}}
\vspace{0pt}\textbf{{\large A}} &\vspace{0pt}\includegraphics[width=\linewidth]{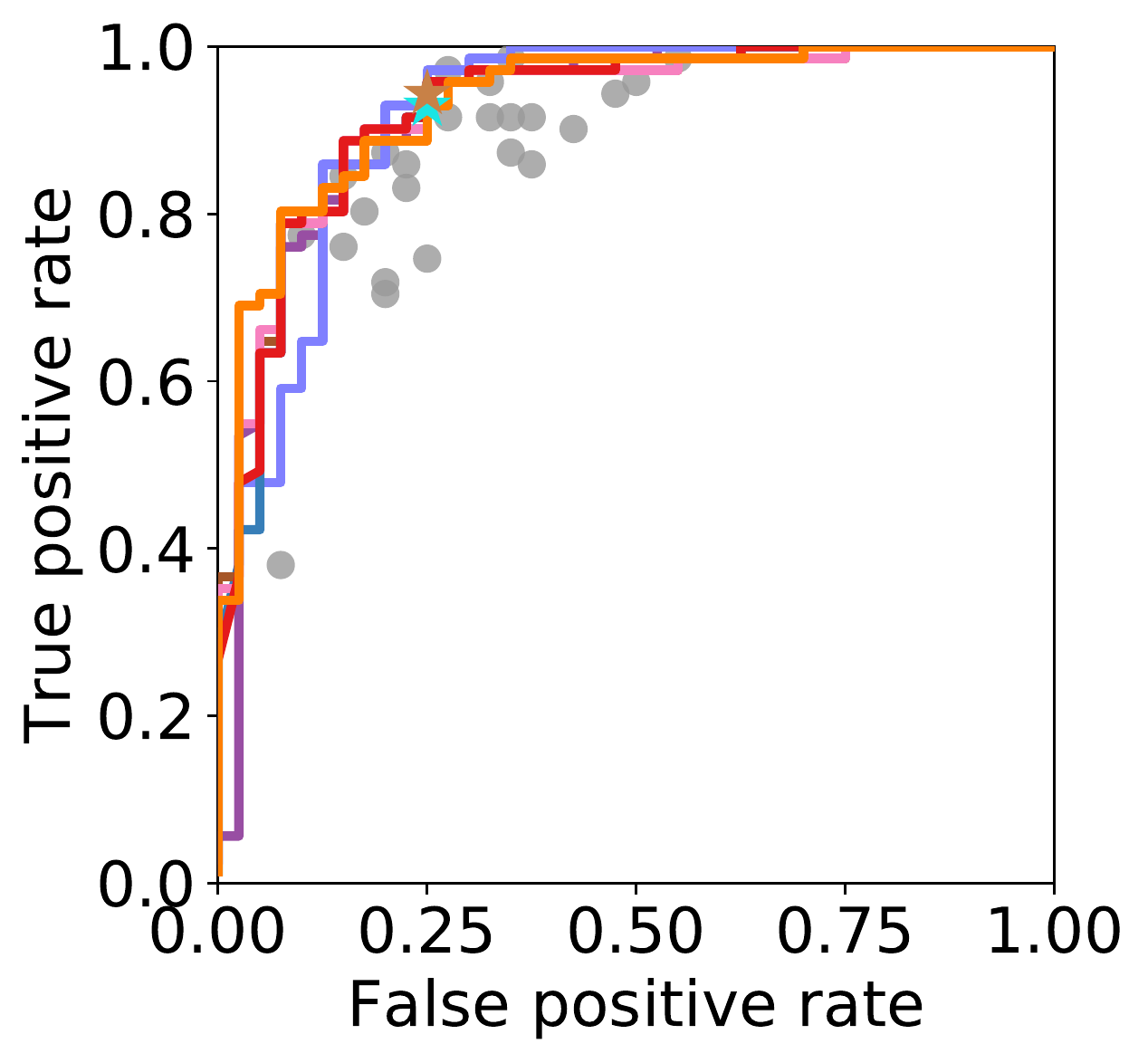}
&\vspace{0pt}\textbf{{\large B}} &\vspace{0pt}\includegraphics[width=\linewidth]{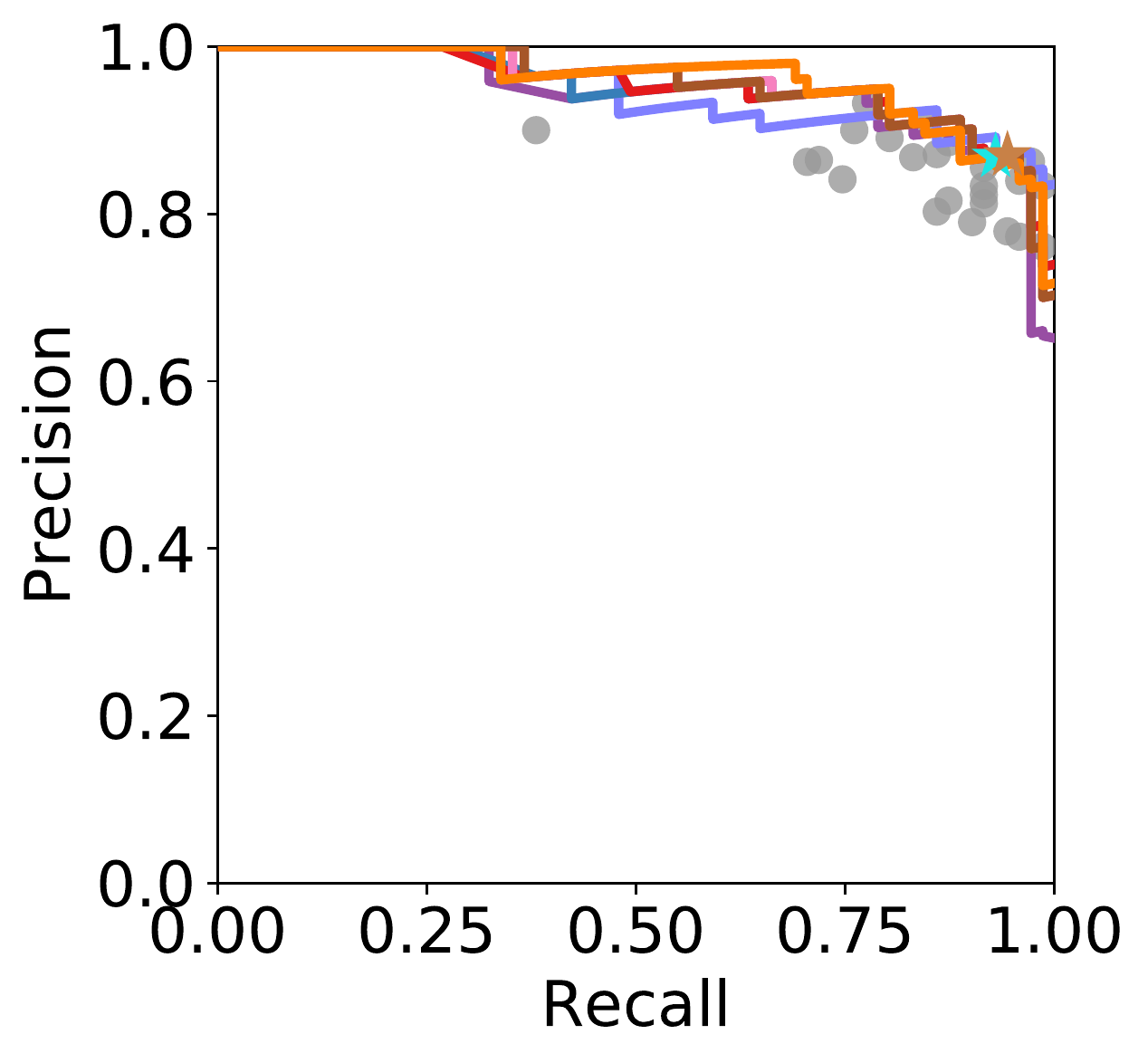}
&\vspace{0pt}\includegraphics[width=\linewidth]{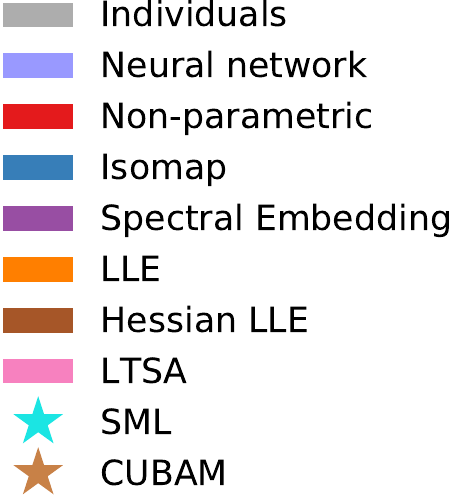}\\
\vspace{0pt}\textbf{{\large C}} &\vspace{0pt}\includegraphics[width=\linewidth]{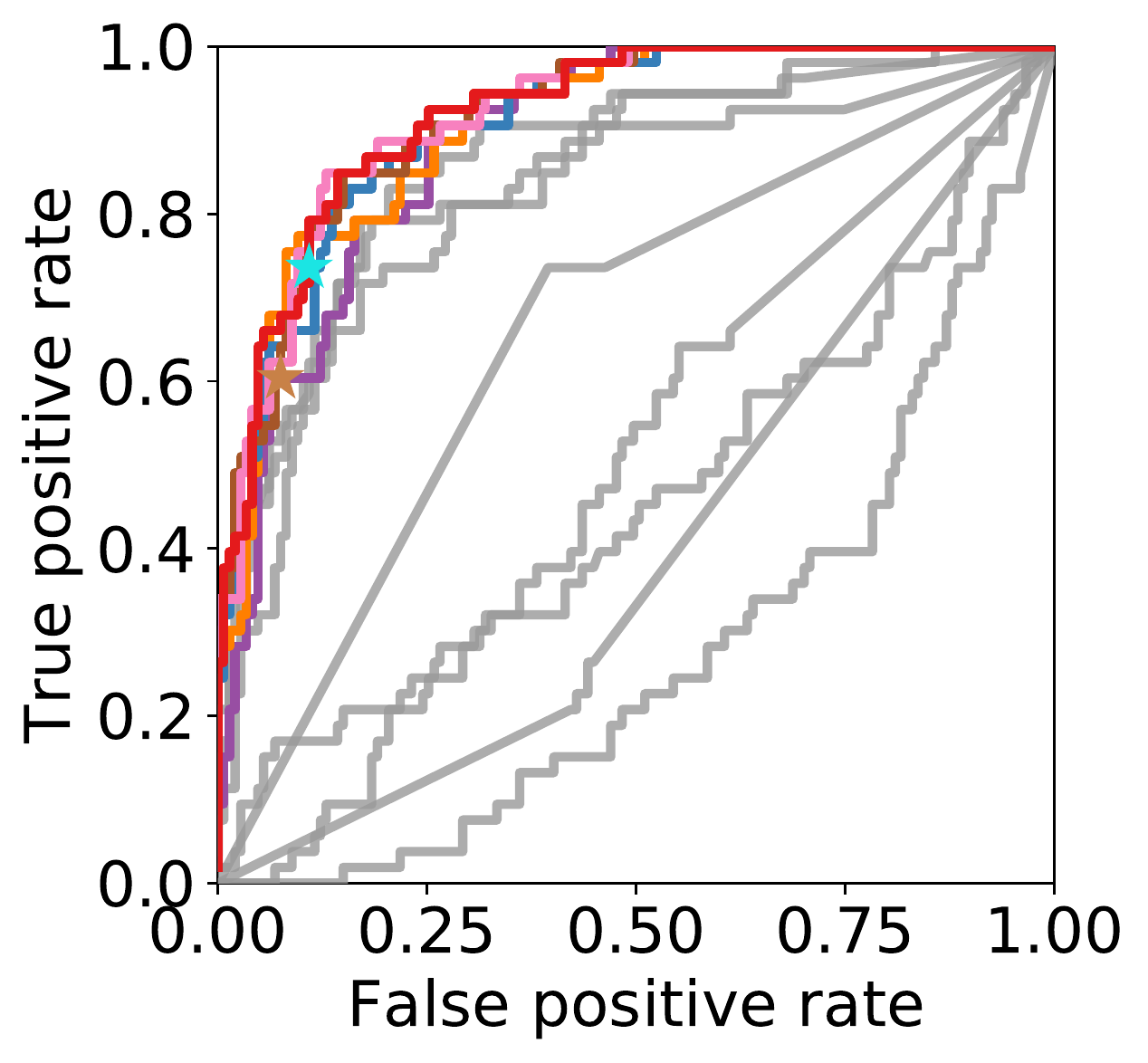}
&\vspace{0pt}\textbf{{\large D}} &\vspace{0pt}\includegraphics[width=\linewidth]{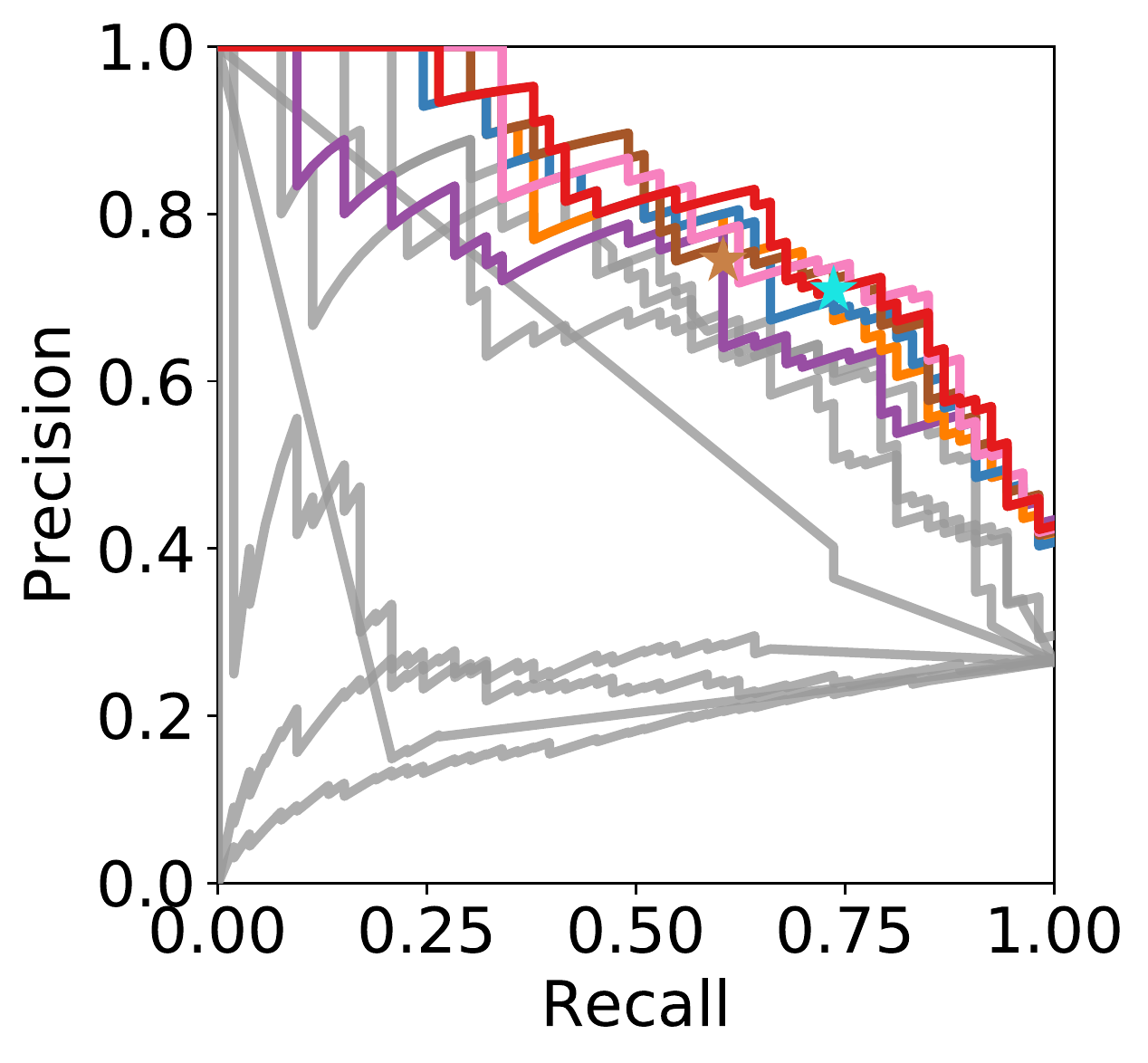}
&\vspace{0pt}\footnotesize
Non-parametric:
\begin{tabular}{ll}
\circled{A}&Mean\vspace{0.5ex}\\
\circled{E}&Median\vspace{0.5ex}\\
\circled{F}&Factor analysis\vspace{0.5ex}\\
\circled{M}&MDS\vspace{0.5ex}\\
\circled{P}&PCA
\end{tabular}
\end{tabular}
\\
\begin{tabular}{p{0em}p{0.45\linewidth}p{0em}p{0.45\linewidth}}
\vspace{0pt}\textbf{{\large E}} &\vspace{0pt}\includegraphics[width=\linewidth]{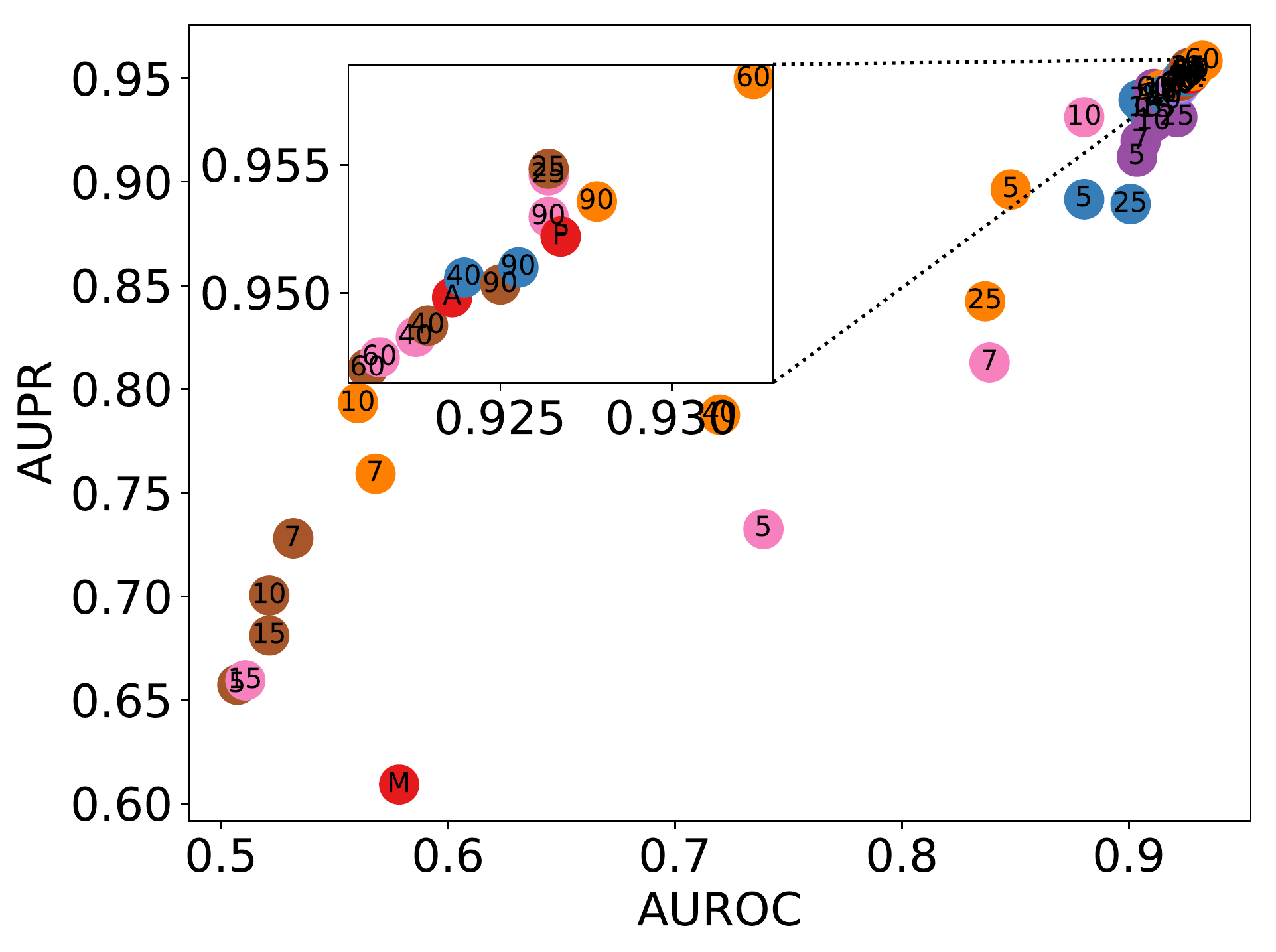}
&\vspace{0pt}\textbf{{\large F}} &\vspace{0pt}\includegraphics[width=\linewidth]{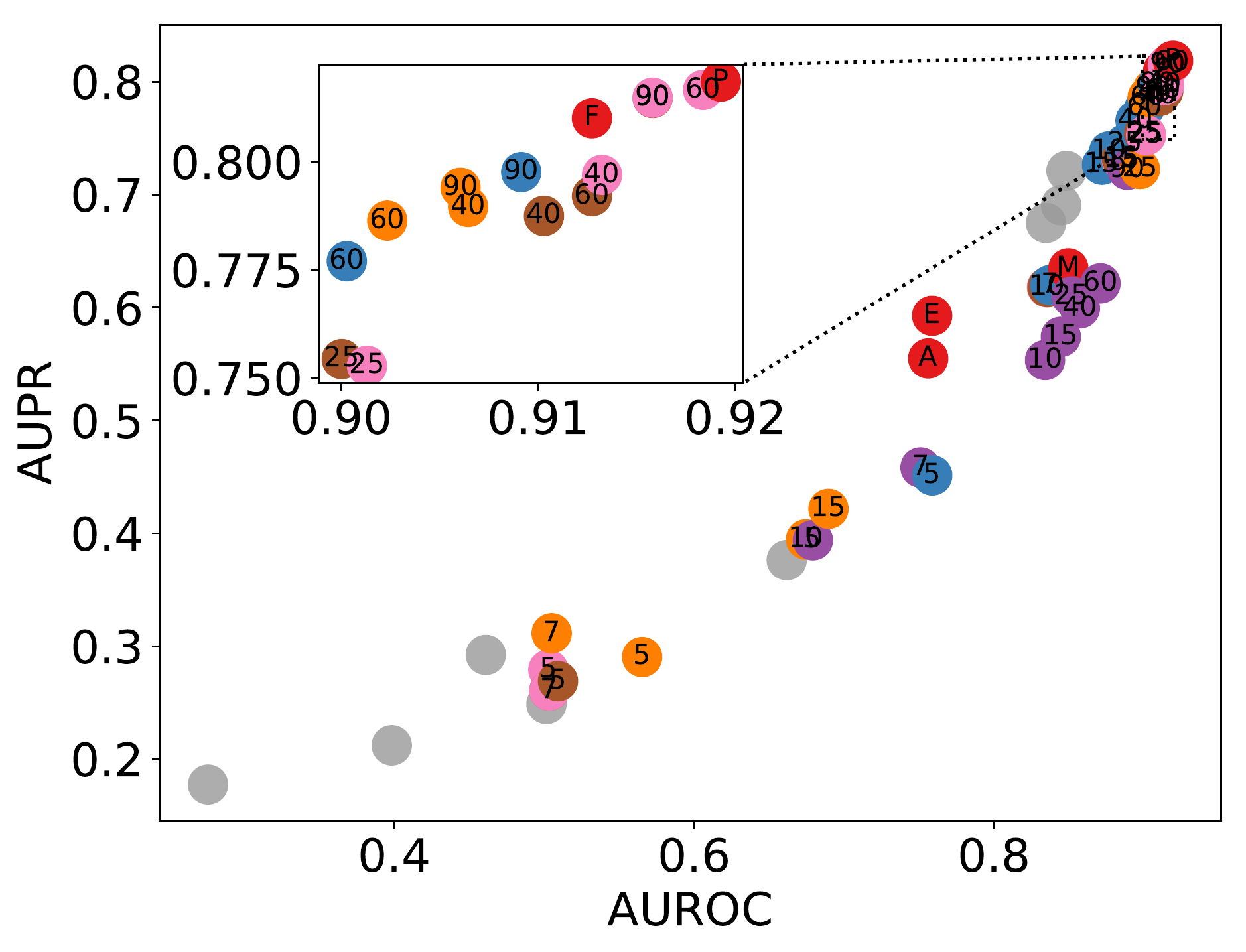}
\end{tabular}
\end{center}
\caption{\textbf{Dimension reduction methods outperformed most or all individuals and existing crowd wisdoms by accounting for confidence information.} (\textbf{A, B, C, D}) ROC (\textbf{A, C}) and Precision-Recall (\textbf{B, D}) curves and plots of individual responses, a deep neural network, existing crowd wisdoms, and selected dimension reduction methods for skin cancer classification (\textbf{A, B}) and the DREAM2 challenge (\textbf{C, D}). The best parameter (in \textbf{E} or \textbf{F}) was selected according to AUROC (\textbf{A, C}) or AUPR (\textbf{B, D}). PCA is selected for non-parametric dimension reduction. SML and CUBAM only accept and output binary responses (\refm). (\textbf{E, F}) AUROC and AUPR from individual responses, dimension reduction, existing crowd wisdom methods, and a deep neural network for skin cancer classification (\textbf{E}) and the DREAM2 challenge (\textbf{F}). The top-right 15 predictions are magnified in the inset. Numbers indicate the number of nearest neighbors.\label{fig-2}}
\end{figure}

To first evaluate dimension reduction methods on binary responses, we envisioned an algorithm-assisted diagnostic committee of 24 dermatologists whose skin cancer classifications are known for 111 dermoscopy images (\cite{esteva_dermatologist-level_2017}). As a comparison, we applied principal component analysis (PCA), factor analysis (FA), multi-dimensional scaling (MDS), locally linear embedding (LLE), Hessian LLE, local tangent space alignment (LTSA), Isomap, and spectral embedding to estimate the class probability ranking from the individual classifications (\refm). PCA and FA were superior to most dermatologists and were among the top crowd wisdoms. Nearest neighbor based methods were not significantly more accurate than PCA, but instead converged towards PCA at large numbers of neighbors, suggesting no significant nonlinearity (\refig{2}\textbf{E}). PCA and FA offered continuous confidence levels which reduced to state-of-the-art binary crowd wisdom solutions from SML (\cite{parisi_ranking_2014}) and CUBAM (\cite{welinder_multidimensional_2010}) at certain thresholds (\refig{2}\textbf{AB}, \refsfig{curves37}\textbf{AB}). Interestingly, more than 15 crowd wisdoms had better classification performance than a deep neural network trained on 130k clinical images (\refig{2}\textbf{ABE}, \refstab{auc}, \cite{esteva_dermatologist-level_2017}). This demonstrates the cutting-edge efficacy from dimension reduction on the binary crowd wisdom task.

To test whether continuous confidence information can improve accuracy, we applied the same dimension reduction methods on the DREAM2 dataset, as well as on their perfectly binarized yes/no responses (\refm). PCA on continuous confidence levels was more accurate than SML and CUBAM on binarized responses (\refig{2}\textbf{CD}, \refsfig{curves20}). Performance differences between crowd wisdoms were in agreement with the skin cancer classification data, except that mean and median --- often the default crowd wisdom method for continuous data (\cite{marbach_wisdom_2012}) --- could not account for worse-than-random individuals (\refig{2}\textbf{CDF}, \refsfig{curves37}\textbf{CD}, \refsfig{corr3}). Many dimension reduction methods, including PCA and Isomap, outperformed every team. Dimension reduction provided reliable and superior crowd wisdom from confidence information without knowing the true class distribution.

\begin{figure}
\begin{center}
\begin{tabular}{p{0em}p{0.8\linewidth}}
\vspace{0pt}\textbf{{\large A}}&\vspace{0pt}\includegraphics[width=\linewidth]{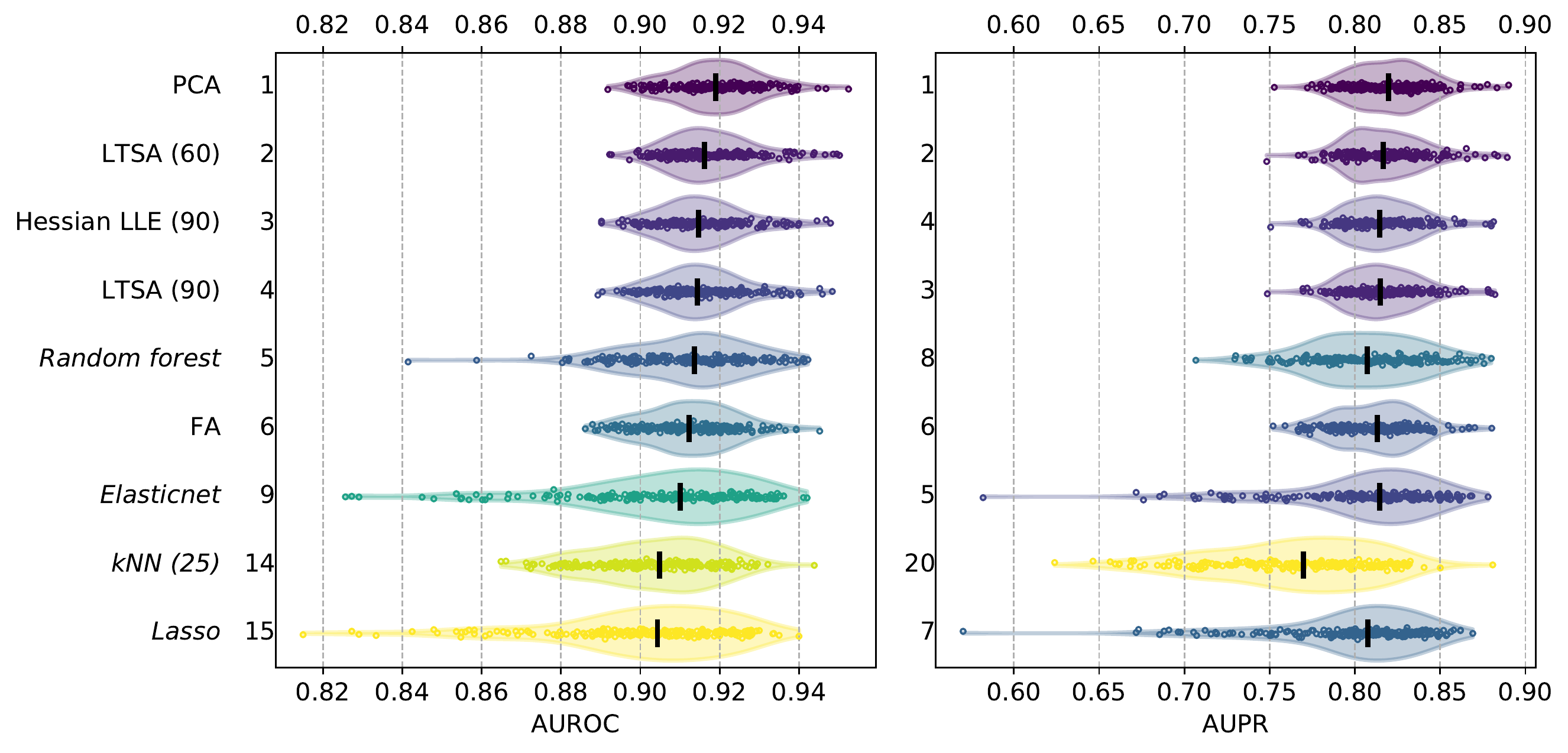}
\\
\vspace{0pt}\textbf{{\large B}}&\vspace{0pt}\includegraphics[width=\linewidth]{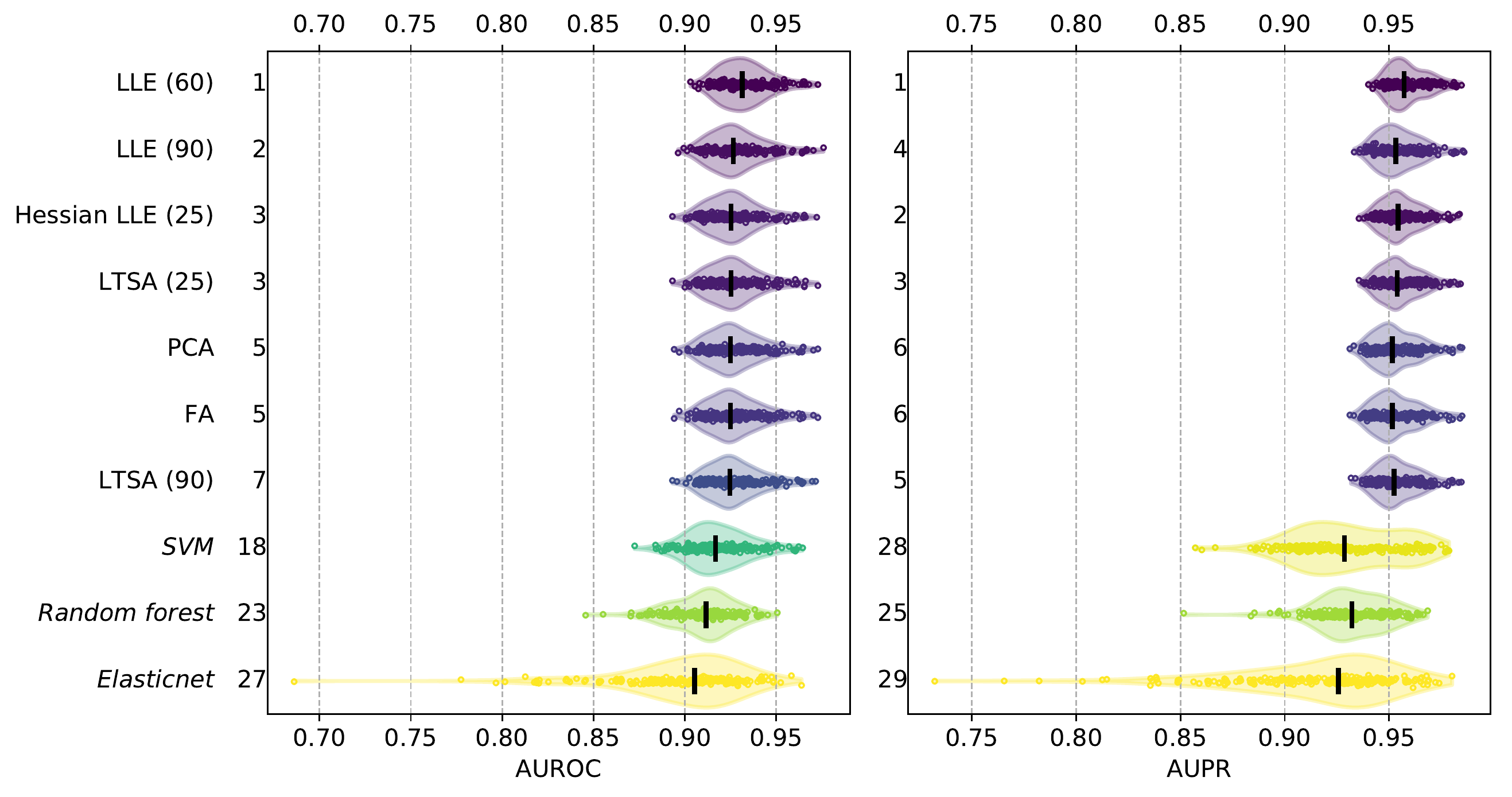}
\end{tabular}
\end{center}
\caption{\textbf{Crowd wisdom outperformed supervised learning in cross-validation.} Empirical distributions and medians of AUROC (left) and AUPR (right) of top crowd wisdom and supervised learning methods in 200 cross-validations with 25\% random partition of training data are shown for the DREAM2 (\textbf{A}) and skin cancer (\textbf{B}) datasets. Method names include the numbers of nearest neighbors in brackets, and are \textit{italicized} for supervised classifiers. Numbers next to the frames represent rankings of the methods in terms of median AUROC or AUPR among all 66(\textbf{A})/64(\textbf{B}) methods. Colors reflect methods' relative rankings in performance. Only the top 5 crowd wisdoms and top 3 supervised classifiers in either AUROC and AUPR are shown.\label{fig-3}}
\end{figure}

Knowledge of the ground-truth for a subset of questions may help calibrating response aggregations for the remaining questions. For instance, in daily life we trust people and favor programs that were more accurate historically. To compare calibrated response aggregations against ground-truth-ignorant crowd wisdoms, we cross-validated crowd wisdoms and 8 popular supervised classifiers [including linear, logistic, lasso, and elasticnet regression, linear discriminant analysis (LDA), support vector machine (SVM), kNN, and random forest] that were trained on randomly selected question subsets (\refm). Surprisingly, crowd wisdom had equal or better performance than supervised classifiers for both the DREAM2 and the skin cancer datasets in terms of AUROC and AUPR (\refig{3}, \refsfig{supa3}, \refsfig{supa7}). Supervised classifiers could only reach crowd wisdom's performance with 50\% of training data or more (\refsfig{supa3}, \refsfig{supa7}). Considering that the true answers in practical research questions are largely unknown, unsupervised crowd wisdom outperformed supervised learning by integrating the test dataset to better estimate individual accuracies.

\begin{figure}
\begin{center}
\begin{tabular}{p{0em}p{0.42\linewidth}p{0em}p{0.42\linewidth}}
\vspace{0pt}\textbf{{\large A}}&\vspace{0pt}\includegraphics[width=\linewidth]{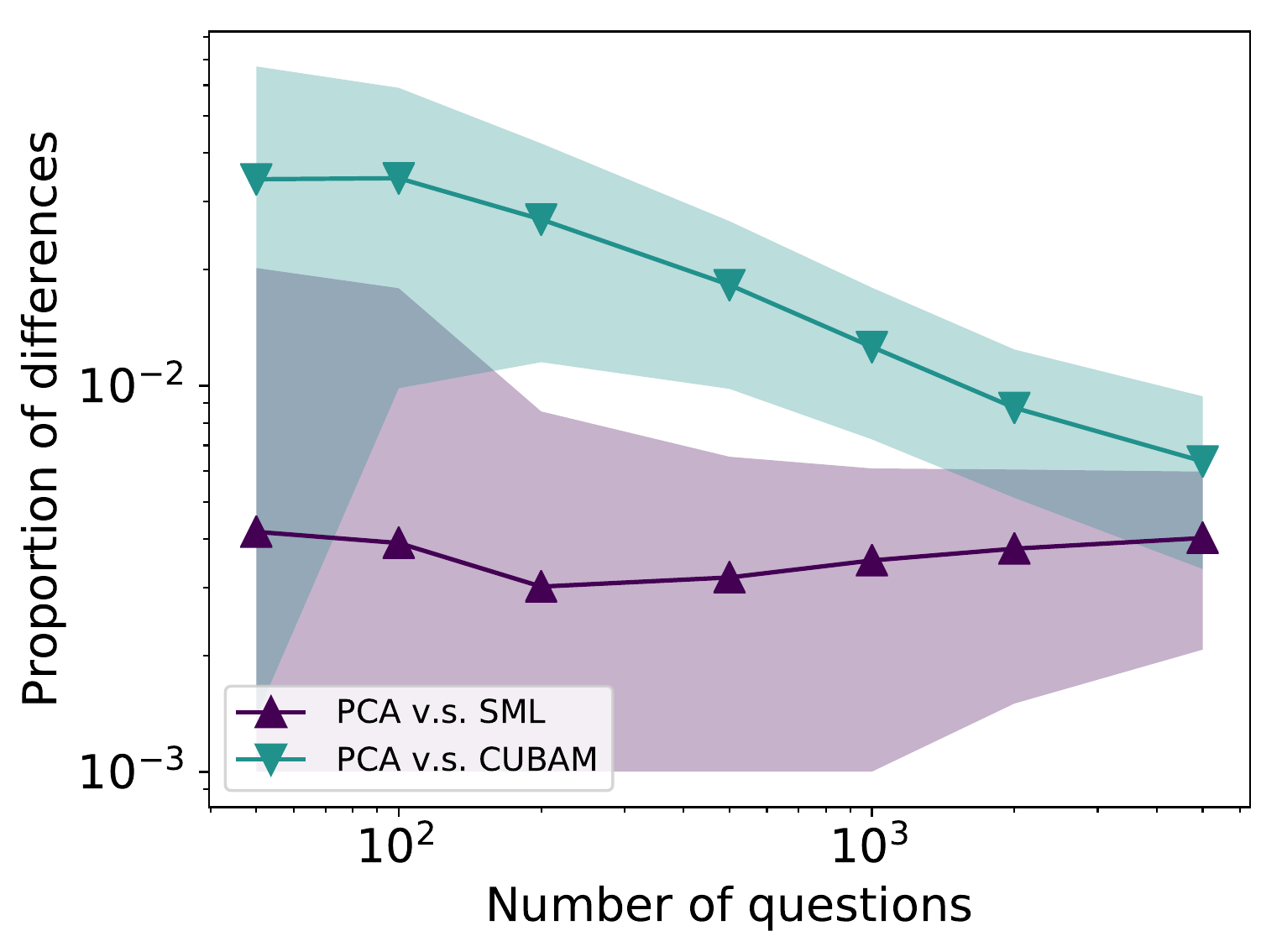}
&\vspace{0pt}\textbf{{\large B}}&\vspace{0pt}\includegraphics[width=\linewidth]{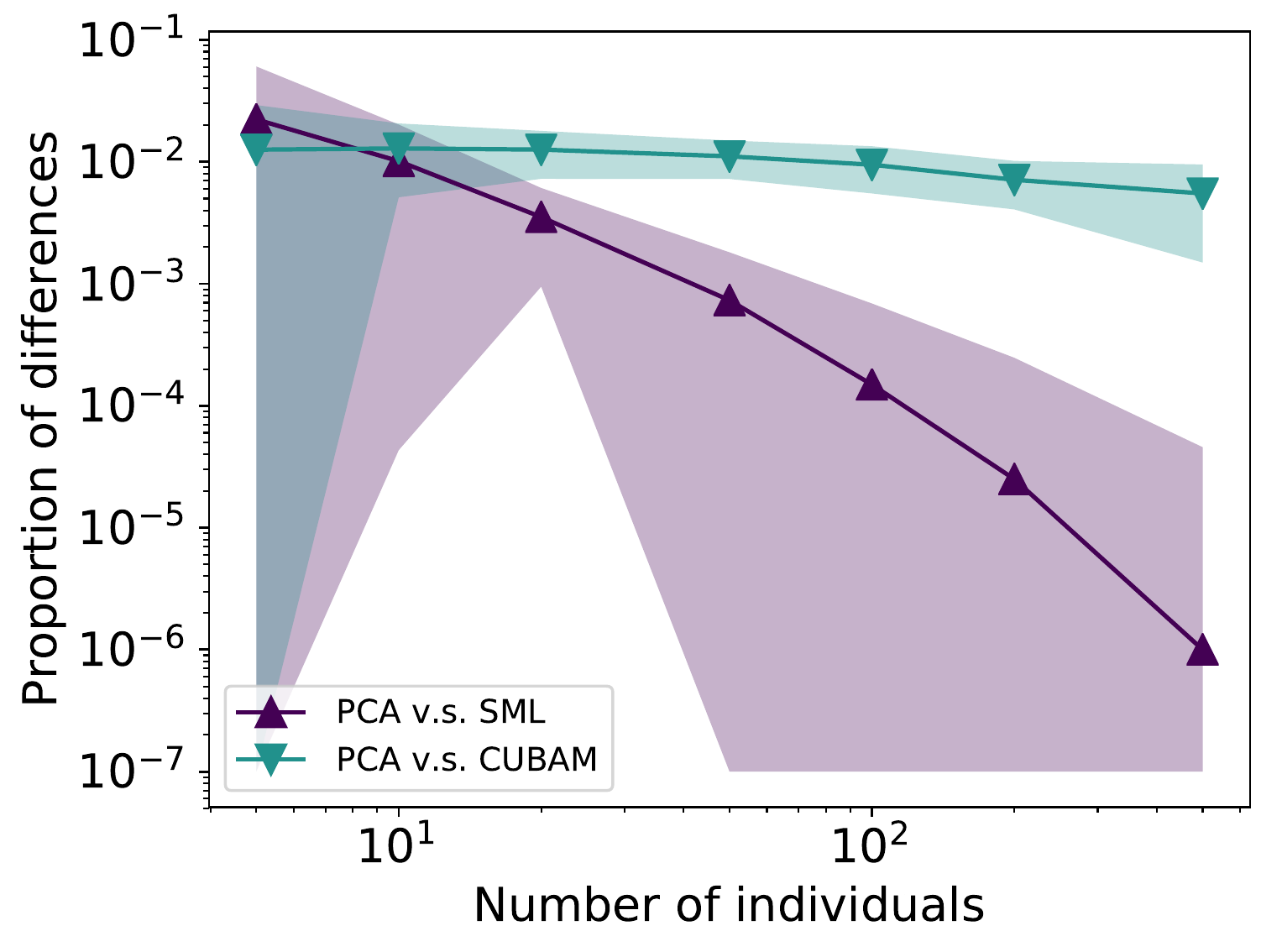}
\end{tabular}
\begin{tabular}{p{0em}p{0.35\linewidth}p{0em}p{0.35\linewidth}p{0.13\linewidth}}
\vspace{0pt}\textbf{{\large C}}&\vspace{0pt}\includegraphics[width=\linewidth]{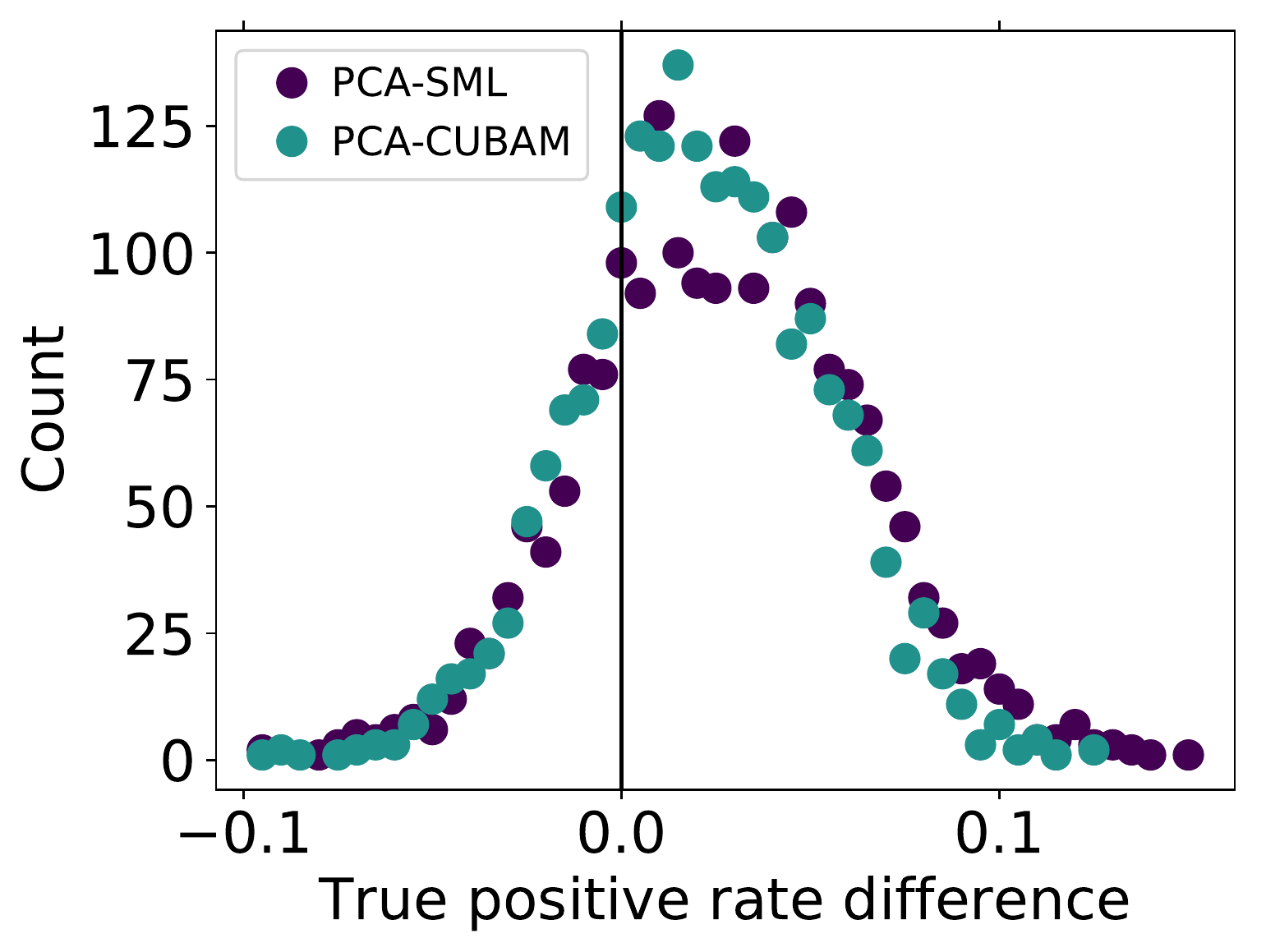}
&\vspace{0pt}\textbf{{\large D}}&\vspace{0pt}\includegraphics[width=\linewidth]{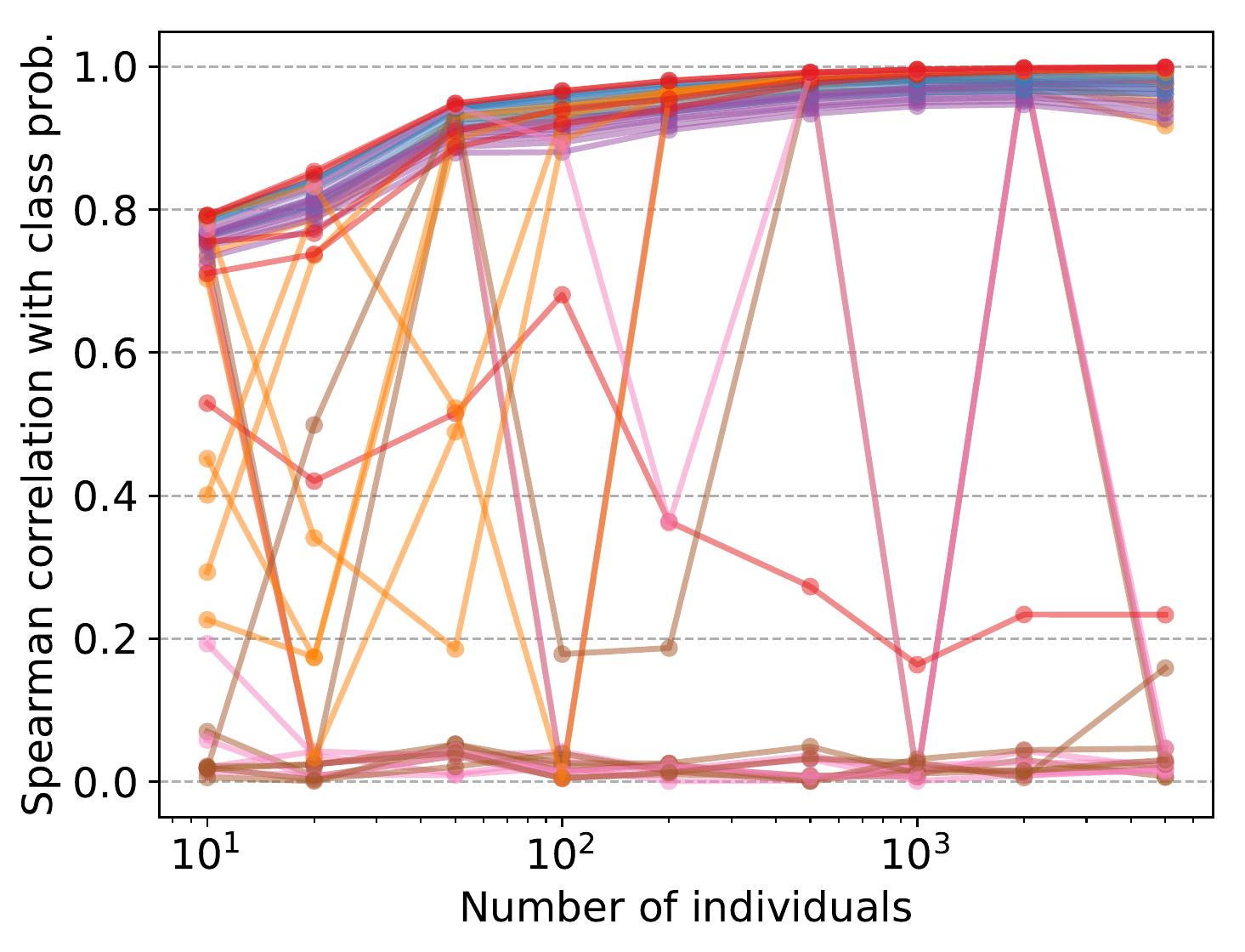}
&\vspace{0pt}\includegraphics[width=\linewidth,trim=0 1cm 0 1cm,clip]{legend.pdf}
\end{tabular}
\end{center}
\caption{\textbf{Dimension reduction methods include SML, outperform previous binary crowd wisdom methods, and recovers the class probability.} (\textbf{A, B}) The proportion of differences in aggregated responses as a function of the number of questions (\textbf{A}) or individuals (\textbf{B}) on simulated binary datasets. Lines and shades represent mean and standard deviation in 2000 random repeats. (\textbf{C}) Empirical distribution of true positive rate differences between PCA and existing binary crowd wisdoms at their respective false positive rate thresholds in 2000 random repeats of simulation 1. (\textbf{D}) Absolute Spearman correlation between each crowd wisdom and the class probability at different numbers of individuals. See \refm{} for details.\label{fig-4}}
\end{figure}

We further interrogated crowd wisdoms in controlled simulations. With 2000 replicated simulations for each parameter set, we found SML to highly correlate with and converge to thresholded PCA as the number of individuals increases (\refig{4}\textbf{AB}, \refm). SML was consequently less sensitive than PCA due to the loss of information, even in perfect binarizations of confidence levels (\refig{4}\textbf{C}, Student's $t$-tests $P<10^{-160}$, \refsfig{density}, \refm). CUBAM was also less sensitive after binarization than PCA. In single simulations (\refm), PCA, FA, Isomap, and LLE converged to perfect class probability predictions as the number of individuals increased (\refig{4}\textbf{D}, \refsfig{converge}, \refsfig{supsimauc}\textbf{B-I}), but LLE based methods were unreliable on noisy datasets (\refsfig{supsimauc}\textbf{JK}, \cite{chang_robust_2006}). Single simulations also reaffirmed our existing conclusions. PCA, FA, and Isomap continued to lead the performances (\refsfig{sim1}, \refsfig{supsimauc}) and crowd wisdom remained superior to supervised classifiers (\refig{sim1}\textbf{B}, \refsfig{supsima5}). Mean and median were again hindered by worse-than-random individuals (\refig{sim1}\textbf{A}, \refsfig{supsimauc}). Overall, PCA and Isomap are more reliable and accurate than other dimension reduction methods and previous wisdom of the crowd methods.

By embedding wisdom of the crowd in unsupervised dimension reduction, we have found that PCA and Isomap are efficient, accurate, consistent, and generic algorithms. Unsupervised dimension reductions obtain superior performances over calibrated crowd wisdoms from supervised classifiers. This study does not consider datasets with missing values (\cite{sheshadri_square:_2013}), strongly correlated errors between individuals (\cite{lorenz_how_2011,jaffe_unsupervised_2016}), or post-crowd-wisdom thresholding. Future research on these problems within the dimension reduction framework may further widen the applications of crowd wisdom.

\cc[More references]

\section*{Methods}
\subsection*{DREAM2 BCL6 Transcription Factor Prediction challenge dataset}
The DREAM2 BCL6 Transcription Factor Prediction Challenge is an open crowd challenge to infer BCL6 gene's transcriptional targets (\cite{comm_gustavo_vogel_20167,noauthor_dream2_nodate,klein_transcriptional_2003}). Participating teams inferred BCL6 targets from gene expression microarray and optional external data, and submited confidence scores for 200 potential target genes. Submissions were evaluated against the gold standard derived from ChIP-on-chip and perturbation experiments, containing 53 BCL6 targets. We had access to submissions from 11 teams, in which 8 were full (without missing predictions) and were used for crowd wisdom.

\subsection*{Skin cancer classification dataset}
Deep neural networks outperformed an average dermatologist in the classification of skin cancer from dermoscopy images (\cite{esteva_dermatologist-level_2017}). Based on dermoscopy images alone, dermatologists were asked whether to biopsy/treat the lesion or to reassure the patient. We obtained 24 dermatologists' responses to 111 biopsy-proven dermoscopy images in which 71 were malignant. We also obtained the predicted confidence scores for these images from the deep neural network in \cite{esteva_dermatologist-level_2017}.

\subsection*{Simulated datasets}
A simulated dataset of $n$ binary (yes/no) questions contains their true classes, the (posterior) class probabilities given all the relevant data for each question as $P_i(Yes\mid data)$, and the responses from $k$ individuals to all $n$ questions as matrix $\mathbf{R}=\{r_{ji}\}$, for $i=1,\dots,n$, $j=1,\dots,k$. Given the desired occurrence frequency of class yes as $P(Yes)$, the dataset needs to contain $nP(Yes)$ questions in class yes and $n(1-P(Yes))$ in class no. We simulated the true classes, class probabilities, and individual responses (\refig{1}\textbf{B}) according to the following steps:
\begin{enumerate}
\item Simulate class probabilities $P(Yes\mid data)\sim B(\beta,\beta)$, where $B$ is the Beta distribution and $\beta$ characterizes the question difficulty given all the data. For each question, set the true class to yes with probability $P(Yes\mid data)$ and no otherwise. Only the first $nP(Yes)$ questions in yes class and the first $n(1-P(Yes))$ questions in no class were retained, merged, and shuffled to form the full list of questions $i=1,\dots,n$. Their class probability $P_i(Yes\mid data)$ and true classes were recorded.
\item Simulate individual responses $\mathbf{R}$. Individual $j$'s response to question $i$ is $r_{ji}\sim N(\alpha_j P_i(Yes\mid data),1)$, where $\alpha_j\sim N(\bar\alpha,\sigma_\alpha^2)$.
\item Normalization was applied (cf below).
\end{enumerate}
The simulation takes 6 parameters: $k$, $n$, $P(Yes)$, $\beta$, $\bar\alpha$, and $\sigma_\alpha$. See \refstab{sim} for parameter values.

\subsection*{Perfect binarization}
To transform confidence level responses to binary (yes/no) responses, we chose the ideal scenario for existing binary crowd wisdom methods, by assuming that each individual knows the true total number of yes responses. Consequently, each individual will select that same number of their most confident predictions as yes, and the rest as no. Ties at the yes/no boundary are selected at random.

\subsection*{Normalization}
We normalized raw answers from multiple individuals to multiple questions before applying crowd wisdom or supervised learning (in cross validation). For continuous datasets, we first converted raw answers into rankings, separately for each individual and with ties averaged. Then, for all datasets, we shifted the raw or rank-converted values to zero mean and scaled them to unit variance, separately for each individual.

\subsection*{Dimension reduction as wisdom of the crowd}
From the python package \textit{scikit-learn}, we applied the following dimension reduction methods for crowd wisdom: \textit{TruncatedSVD} (as PCA) and \textit{FactorAnalysis} in \textit{sklearn.decomposition}, and \textit{LocallyLinearEmbedding} (with methods standard, hessian, and ltsa), \textit{Isomap}, and \textit{SpectralEmbedding} in \textit{sklearn.manifold}. Nearest-neighbor based methods took 5, 7, 10, 15, 25, 40, 60, and 90 neighbors. We also included mean and median as simple statistics for crowd wisdom.

\subsection*{Evaluation metrics}
We used the Receiver Operating Characteristic (ROC) and Precision-Recall curves, as well as their areas under the curves (AUROC and AUPR) as evaluation metrics. To tackle the sign indeterminacy from dimension reduction, we always computed these metrics twice, on the original output and on its negative, and selected the one with a larger area under the curve for comparison. For fair comparison, the same procedure was applied on supervised learning methods. In practice, sign indeterminancy can be solved by assuming more than half of the individuals have better-than-random responses, and then aligning crowd wisdom with the majority of the crowd.

\subsection*{Supervised classifiers}
From the python package \textit{scikit-learn}, we applied the following supervised classifiers: \textit{LinearRegression}, \textit{ElasticNetCV}, \textit{LassoCV}, and \textit{LogisticRegression} in \textit{sklearn.linear\_model}, \textit{LinearDiscriminantAnalysis} in \textit{sklearn.discriminant\_analysis}, \textit{RandomForestClassifier} in \textit{sklearn.ensemble}, and \textit{KNeighborsClassifier} in \textit{sklearn.neighbors} with 5, 7, 10, 15, 25, 40, 60, and 90 neighbors.

\subsection*{Method comparison in cross validation}
To compare crowd wisdom and supervised classifiers, we randomly split each dataset into a training set (containing 10, 20, 25, 40, 60, 80, or 90 percent of all samples) and a test set (for the rest), using \textit{sklearn.model\_selection.StratifiedShuffleSplit} and requiring the number of questions to be larger than that of individuals in the training set. Supervised classifiers were trained on individual predictions against ground-truths in the training set, and then predicted for the test set. For crowd wisdom, we performed crowd wisdom on the full data (not using ground-truth) and then extracted predictions for the test set. Evaluation metrics were computed for every random split. The random split was repeated 200 times per split ratio per dataset.

\subsection*{Method comparison on binarized data}
With a given parameter set for simulation, we performed 2000 replicated simulations with different random seeds. For each replicate, the ROC curve for PCA and the FPR and TPR for SML and CUBAM were computed. The ROC quantiles were computed as the quantiles of TPR at every FPR level among the 2000 ROCs from replicates. The densities of SML and CUBAM points on ROC were computed with \textit{scipy.stats.gaussian\_kde}. The TPR difference between PCA and SML or CUBAM was computed at SML's or CUBAM's FPR in each replicate.

\subsection*{Proportion of differences between binary crowd wisdoms and PCA}
Simulations 21, 22, 23, 19, 1, 20, 25 and 24, 3, 1, 4, 5, 6, 7 were used respectively for the comparisons of differences as the numbers of questions and individuals vary. Each simulation consists of 2000 replications with different random seeds. Within each replication, the binary crowd wisdom (SML or CUBAM) and PCA were first applied on the binary(/binarized) simulated data. Should the AUROC between PC1 of PCA and the binary crowd wisdom be $<0.5$, the signs of PC1 are inverted. PC1 is then thresholded so the largest $N$ entries are positive, in which $N$ is the number of positives from the binary crowd wisdom. The proportion of differences is the number of questions on which the thresholded PC1 and the binary crowd wisdom have different predictions, divided by the total number of questions. The mean and standard deviation were then computed across the 2000 replicates, after excluding the (rare) critical failures or single-valued outputs of the binary crowd wisdom.

\section*{Acknowledgements}
LW would like to thank Gustavo Stolovitzky and Robert Vogel for providing the DREAM2 dataset, and Andre Esteva and Sebastian Thrun for providing the skin cancer dataset. This work is supported by BBSRC (grant numbers BB/P013732/1 and BB/M020053/1).

\bibliographystyle{vancouver}
\bibliography{main,main2}

\newpage
\renewcommand\thefigure{S\arabic{figure}}
\renewcommand\thetable{S\arabic{table}}
\setcounter{figure}{0}
\setcounter{table}{0}

\begin{figure}
\begin{center}
\begin{tabular}{p{0em}p{0.45\linewidth}p{0em}p{0.45\linewidth}}
\vspace{0pt}\textbf{{\large A}} &\vspace{0pt}\includegraphics[width=\linewidth]{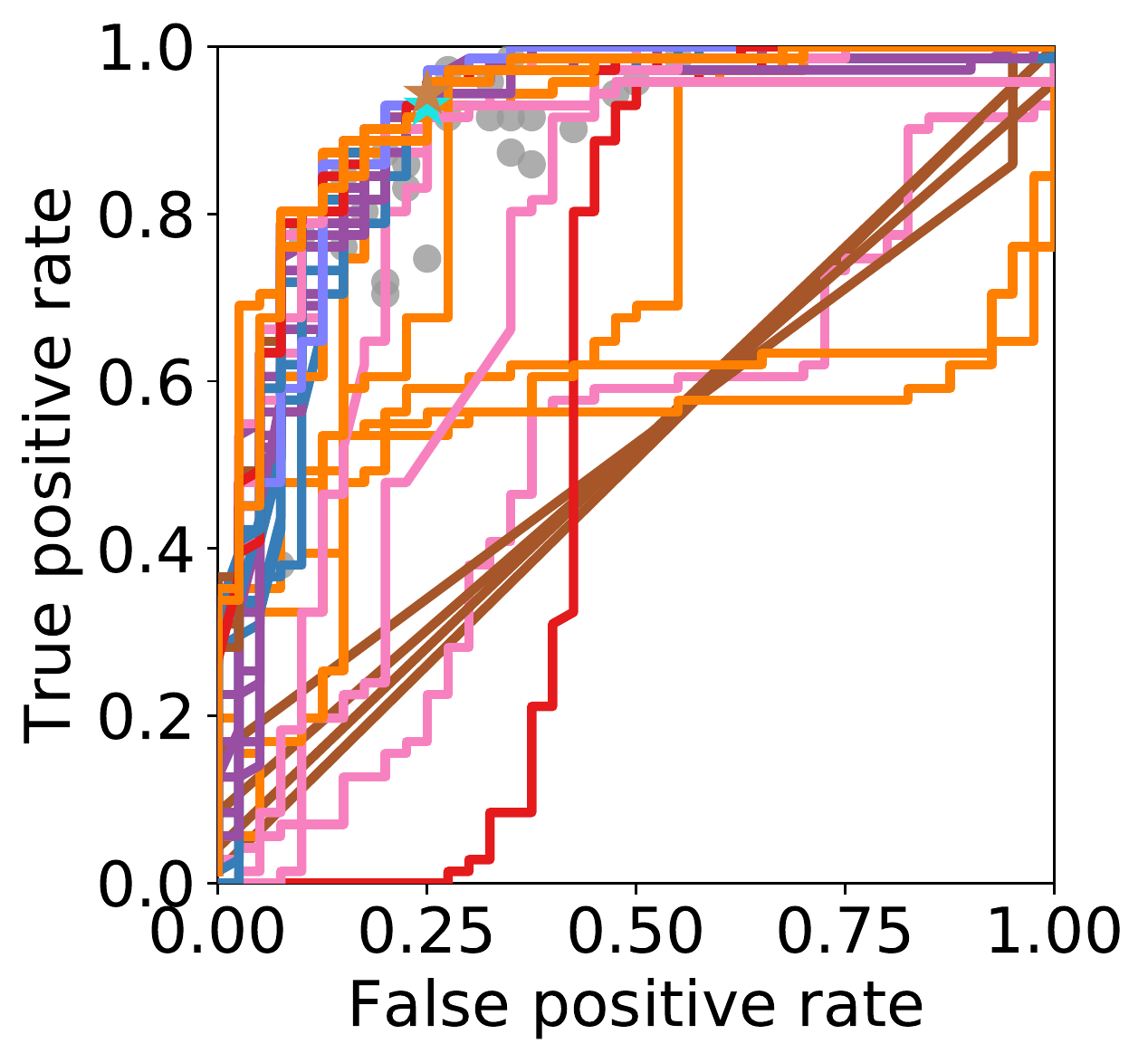}
&\vspace{0pt}\textbf{{\large B}} &\vspace{0pt}\includegraphics[width=\linewidth]{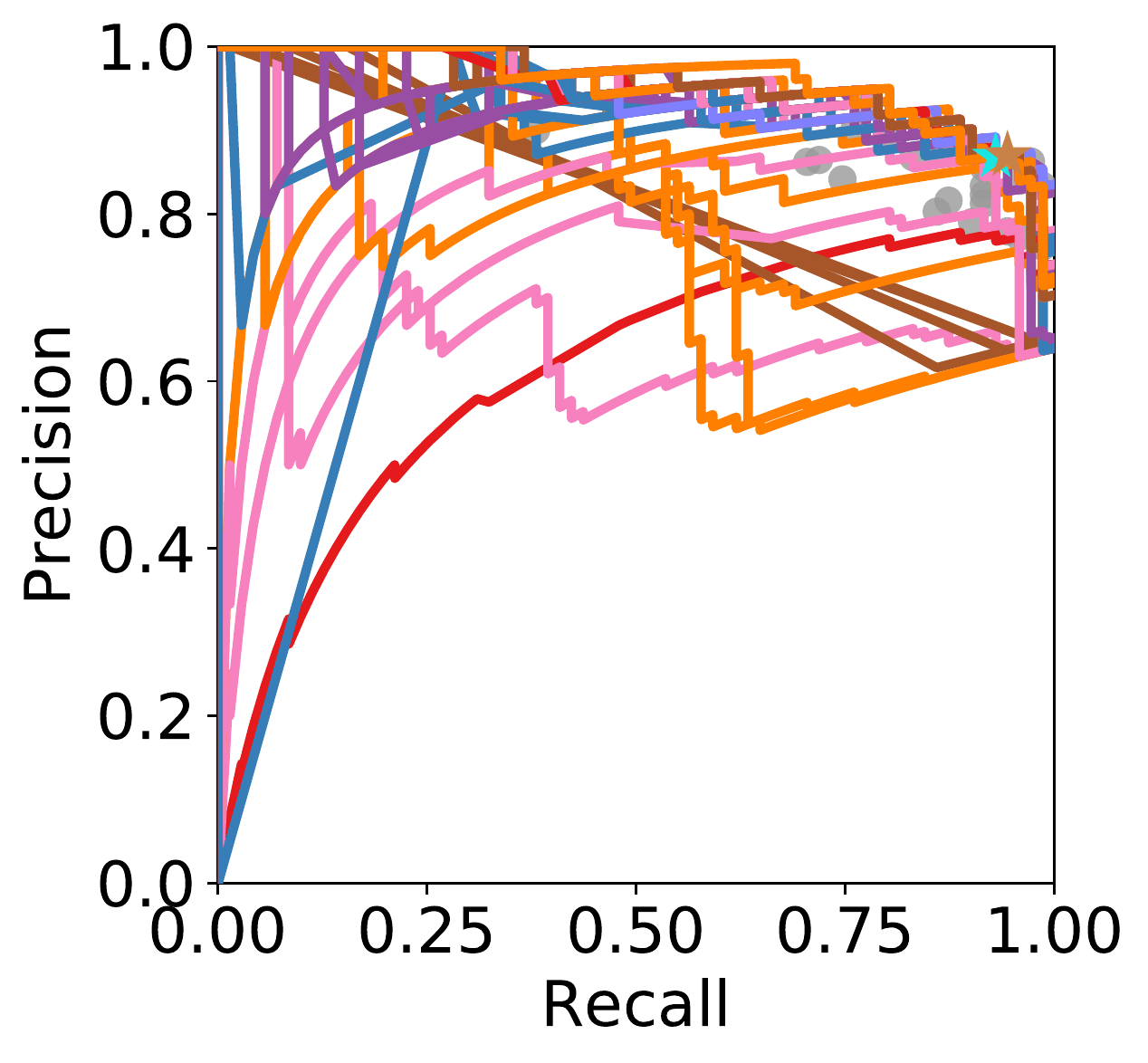}\\
\vspace{0pt}\textbf{{\large C}} &\vspace{0pt}\includegraphics[width=\linewidth]{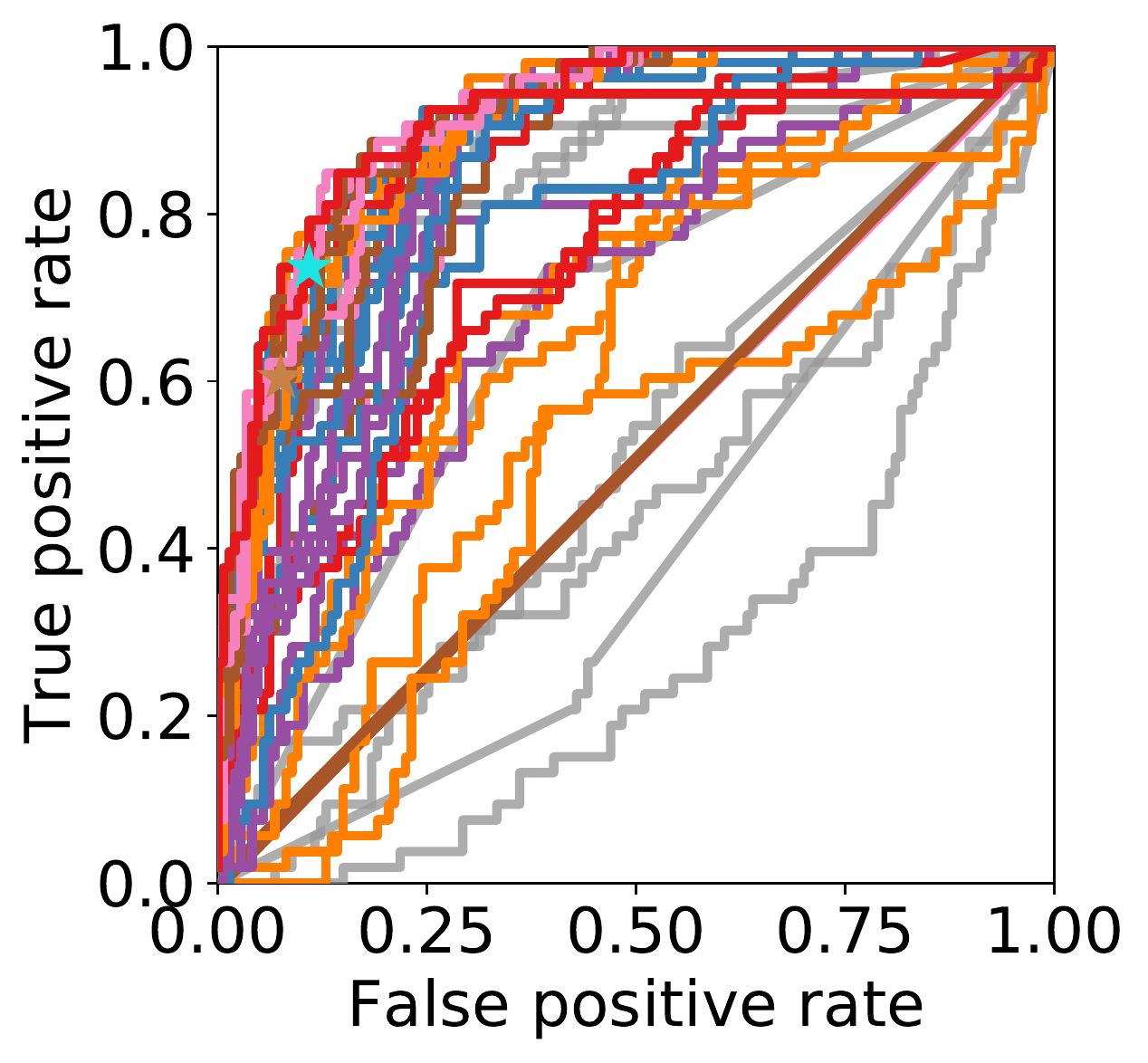}
&\vspace{0pt}\textbf{{\large D}} &\vspace{0pt}\includegraphics[width=\linewidth]{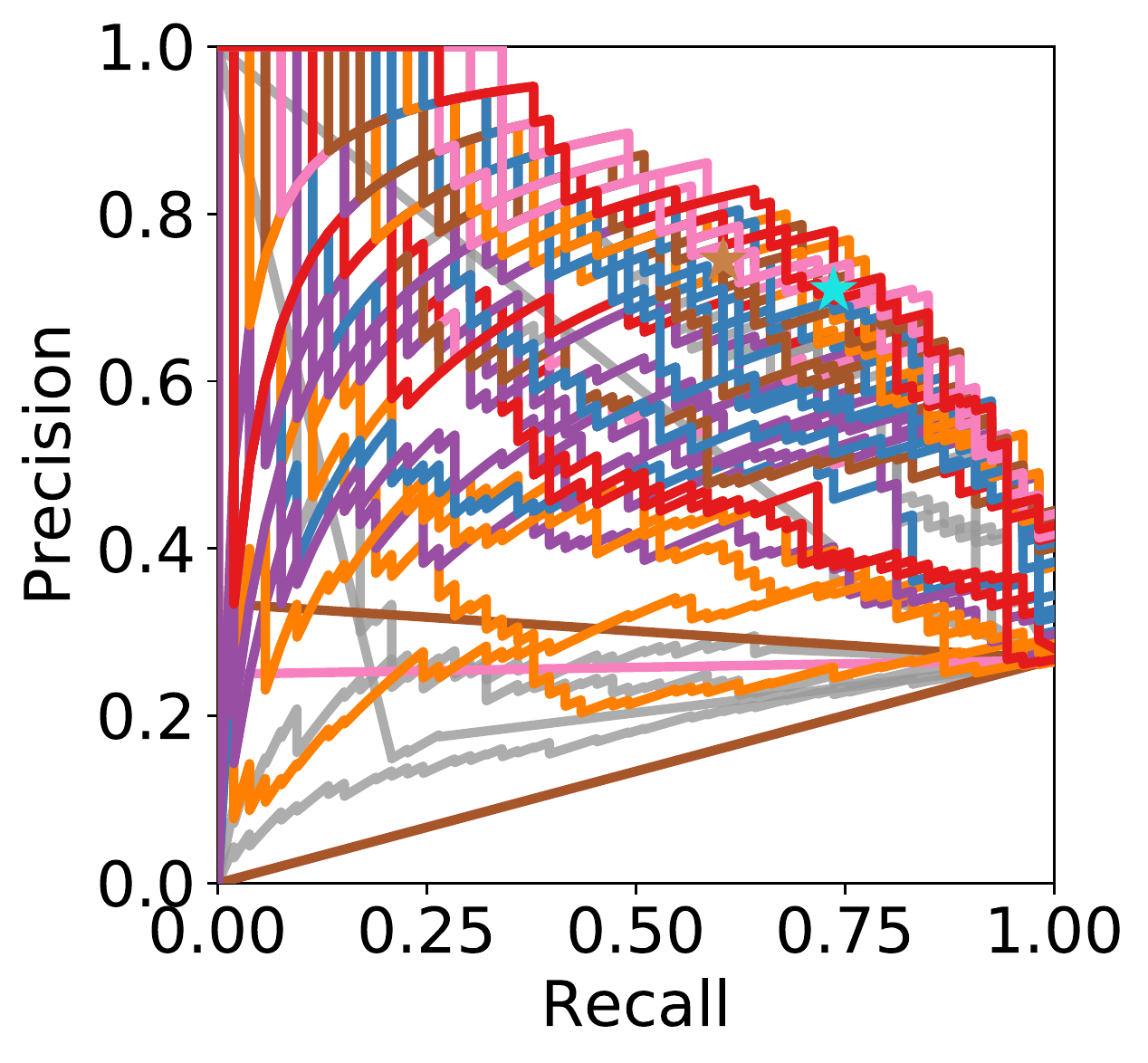}
\end{tabular}
\end{center}
\caption{\textbf{ROC (A, C) and Precision-Recall (B, D) curves for all dimension reductions, existing crowd wisdoms (C, D only), neural network (A, B only) and individual predictions of skin cancer classification (A, B) and DREAM2 challenge (C, D).} For color legend, see \refig{2}.\label{fig-curves37}}
\end{figure}

\begin{figure}
\begin{center}
\begin{tabular}{p{0em}p{0.45\linewidth}}
\vspace{0pt}\textbf{{\large A}} &\vspace{0pt}\includegraphics[width=\linewidth]{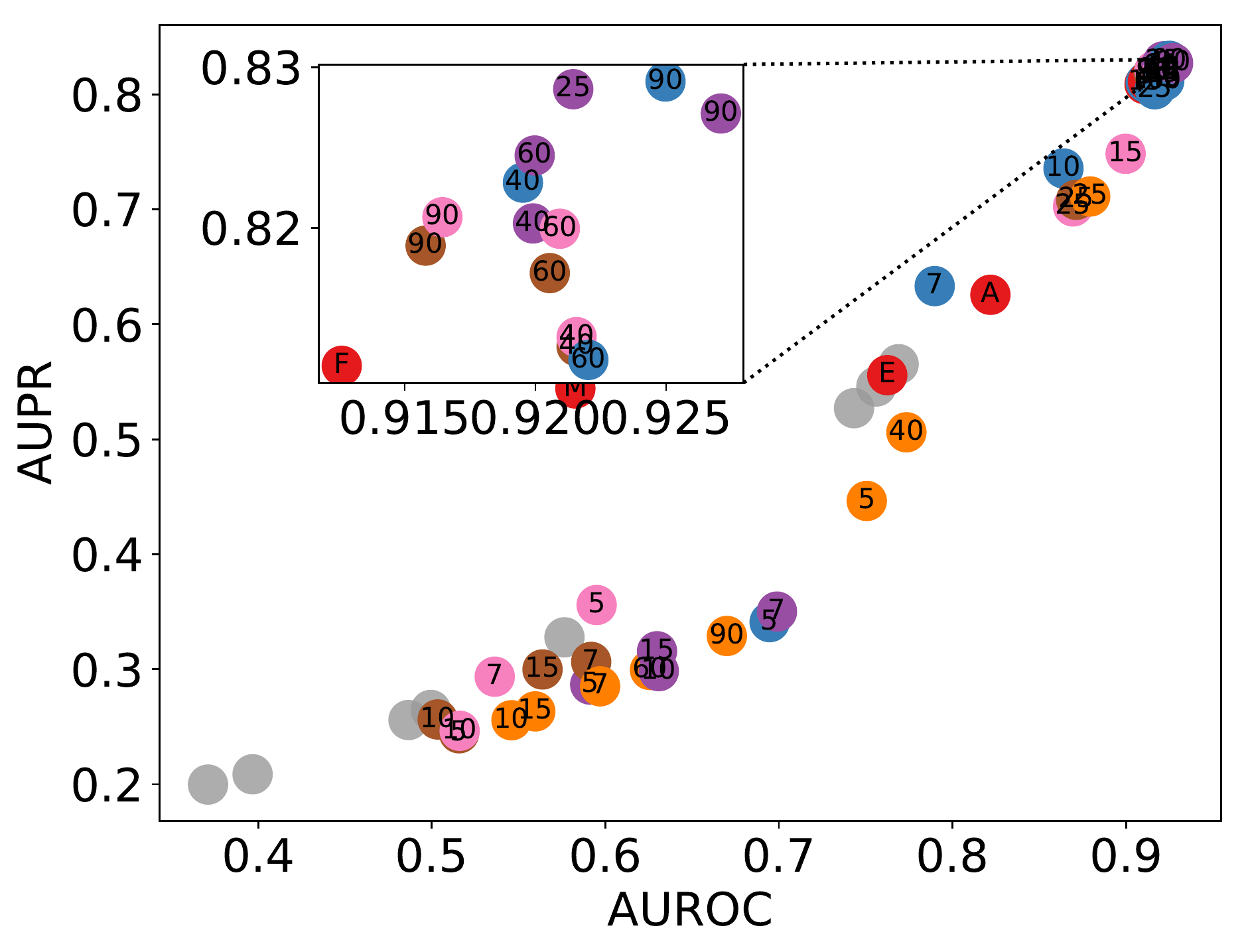}
\end{tabular}
\\
\begin{tabular}{p{0em}p{0.45\linewidth}p{0em}p{0.45\linewidth}}
\vspace{0pt}\textbf{{\large B}} &\vspace{0pt}\includegraphics[width=\linewidth]{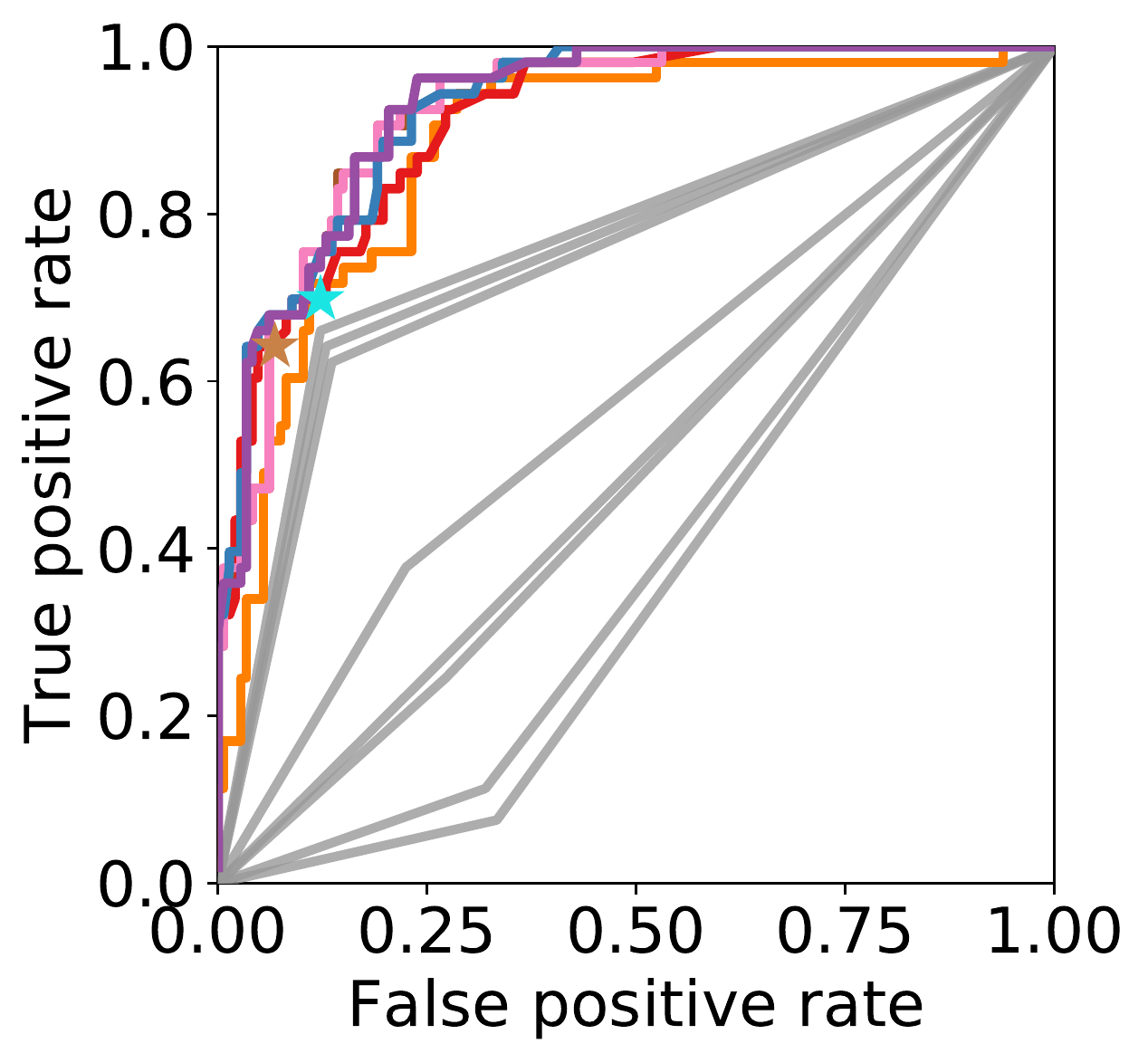}
&\vspace{0pt}\textbf{{\large C}} &\vspace{0pt}\includegraphics[width=\linewidth]{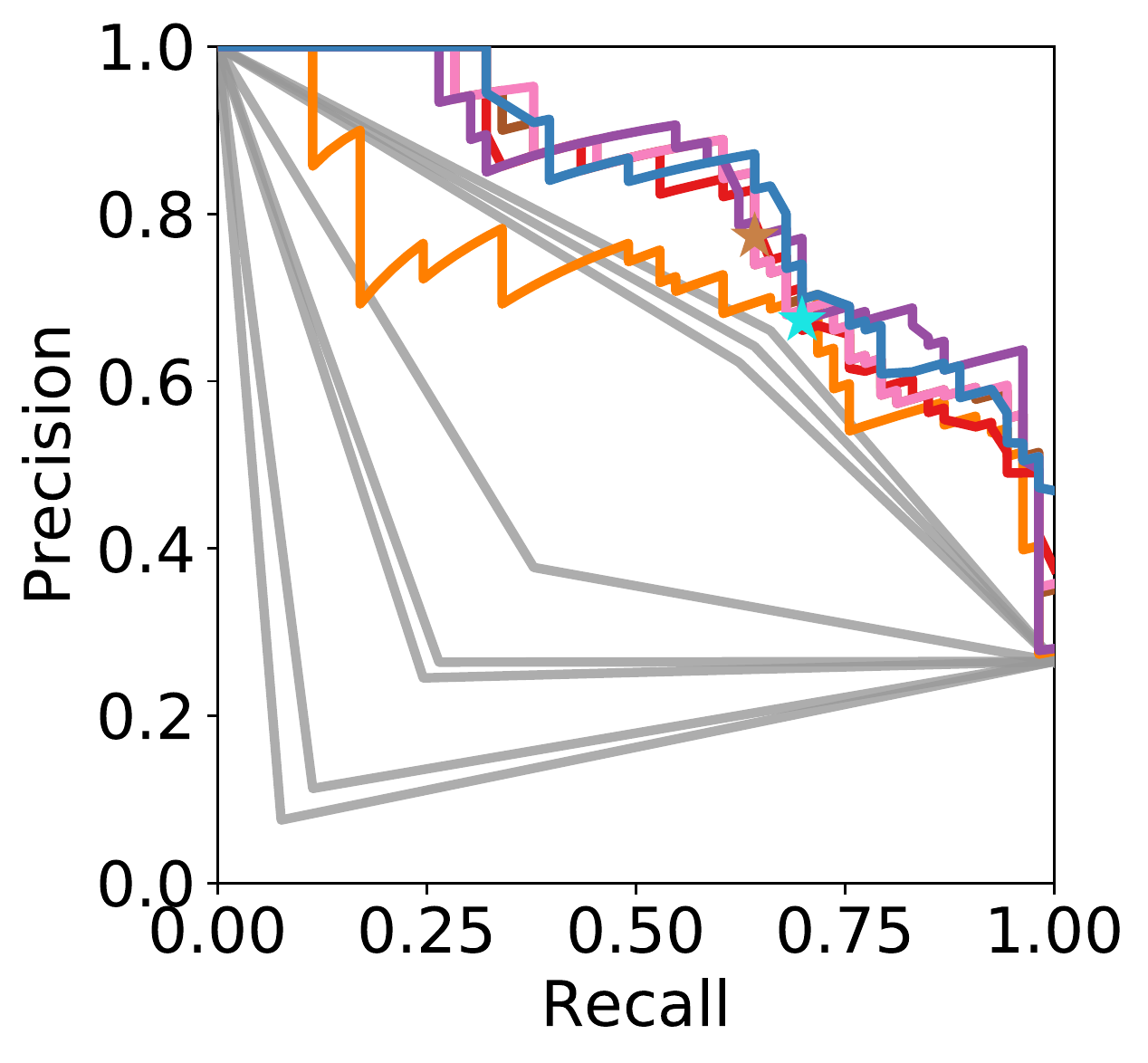}
\end{tabular}
\end{center}
\caption{\textbf{AUPR-AUROC (A), ROC (B) and Precision-Recall (C) for dimension reductions, existing crowd wisdoms, and individual predictions of binarized DREAM2 challenge dataset.} For each parametric dimension reduction, the best parameter (in \textbf{A}) was selected according to AUROC (\textbf{B}) or AUPR (\textbf{C}). PCA was selected for non-parametric dimension reduction. For color legend, see \refig{2}.\label{fig-curves20}}
\end{figure}

\begin{figure}
\begin{center}
\includegraphics[width=0.7\linewidth]{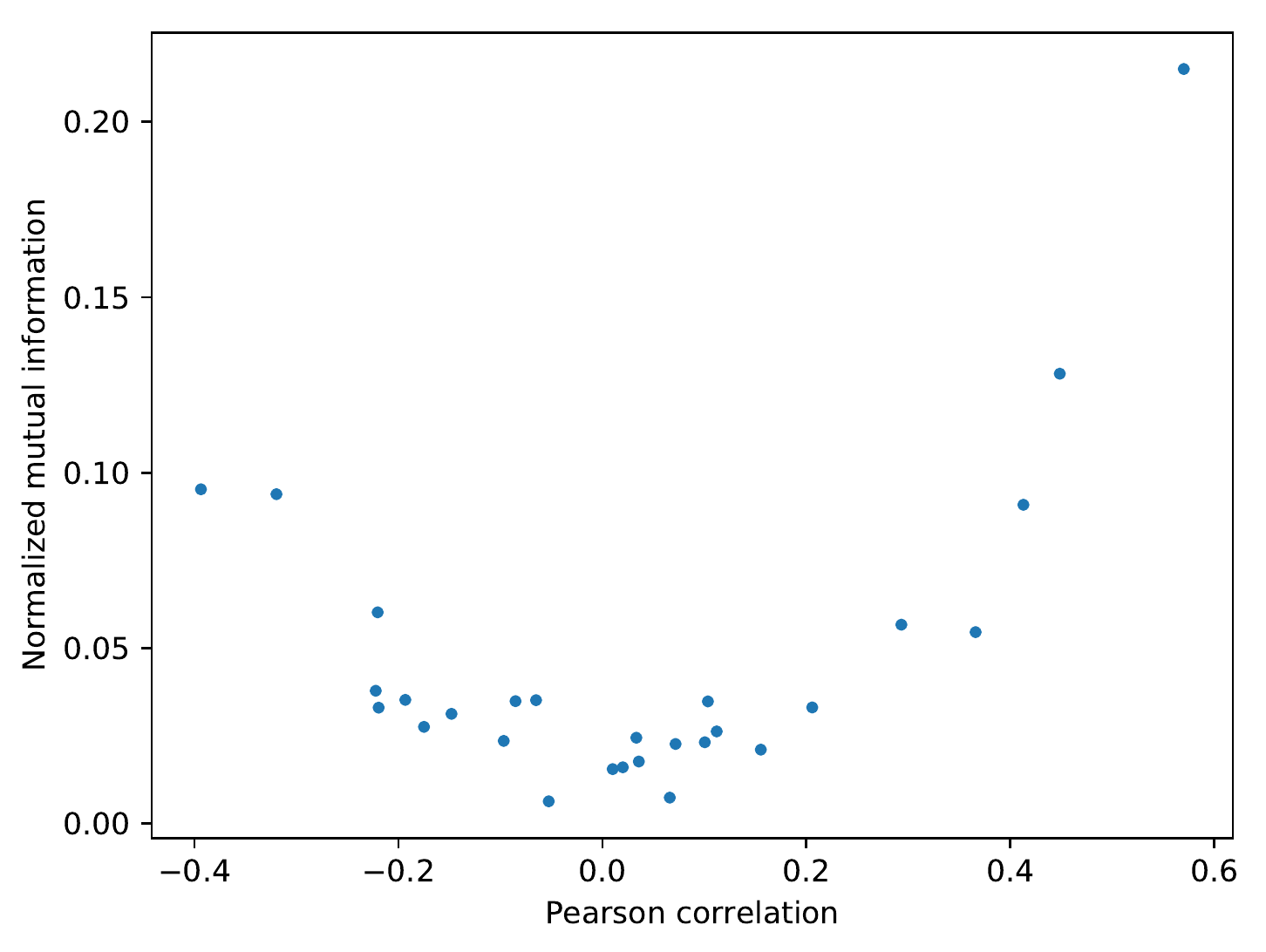}
\end{center}
\caption{\textbf{Scatter plot of Pearson correlation and normalized mutual information between all submissions pairs of the DREAM2 challenge.} Negative Pearson correlations indicate opposite responses from different submissions.\label{fig-corr3}}
\end{figure}

\begin{figure}
\begin{center}
See external file: supa3.pdf
\end{center}
\caption{\textbf{Crowd wisdom outperformed supervised learning in cross-validation.} Empirical distributions and medians of AUROC (left) and AUPR (right) of all crowd wisdom and supervised learning methods in 200 cross-validations with 10\%, 20\%, 25\%, 40\%, 60\%, 80\%, or 90\% (\textbf{A} to \textbf{G}) random partitions of training data are shown for the DREAM2 dataset. Method names include the numbers of nearest neighbors in brackets, and are \textit{italicized} for supervised classifiers. Numbers next to the frames represent rankings of the methods in terms of median AUROC or AUPR. Colors reflect methods' relative rankings in performance.\label{fig-supa3}}
\end{figure}

\begin{figure}
\begin{center}
See external file: supa7.pdf
\end{center}
\caption{\textbf{Crowd wisdom outperformed supervised learning in cross-validation.} Empirical distributions and medians of AUROC (left) and AUPR (right) of all crowd wisdom and supervised learning methods in 200 cross-validations with 25\%, 40\%, 60\%, 80\%, or 90\% (\textbf{A} to \textbf{E}) random partitions of training data are shown for the skin cancer classification dataset. Method names include the numbers of nearest neighbors in brackets, and are \textit{italicized} for supervised classifiers. Numbers next to the frames represent rankings of the methods in terms of median AUROC or AUPR. Colors reflect methods' relative rankings in performance.\label{fig-supa7}}
\end{figure}

\begin{figure}
\begin{center}
\begin{tabular}{p{0em}p{0.6\linewidth}p{0em}p{0.3\linewidth}}
\vspace{0pt}\textbf{{\large A}} &\vspace{0pt}\includegraphics[height=0.55\textwidth]{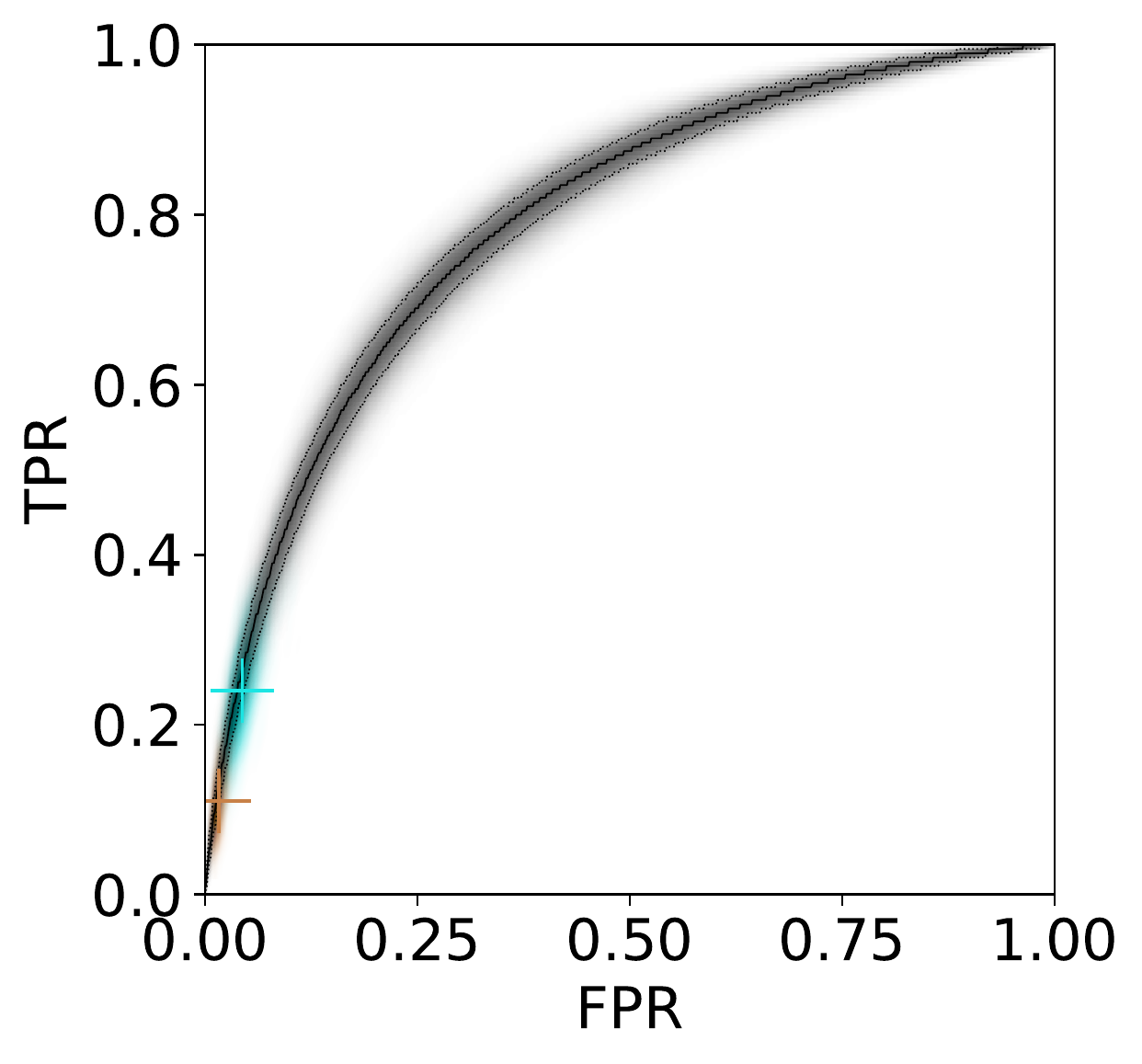}
&\vspace{0pt}\textbf{{\large B}} &\vspace{0pt}\includegraphics[height=0.55\textwidth]{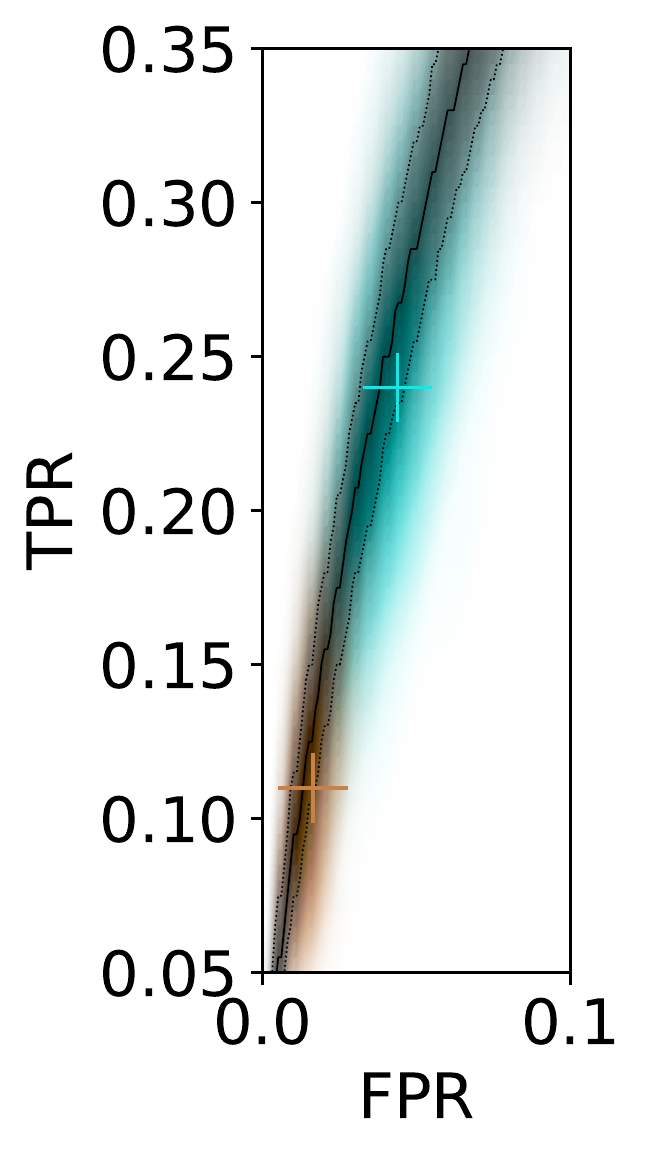}
\end{tabular}
\end{center}
\caption{\textbf{Median and quantiles of ROC curve of PCA (black) and densities of SML (cyan) and CUBAM (brown) ROCs in 2000 random repeats of simulation 1, on full (\textbf{A}) and zoomed-in (\textbf{B}) axes.} Median false positive rate (FPR) and true positive rate (TPR) are shown as `+'. For color legend for SML and CUBAM, see \refig{4}\textbf{C}.\label{fig-density}}
\end{figure}

\begin{figure}
\begin{center}
\includegraphics[width=0.7\linewidth]{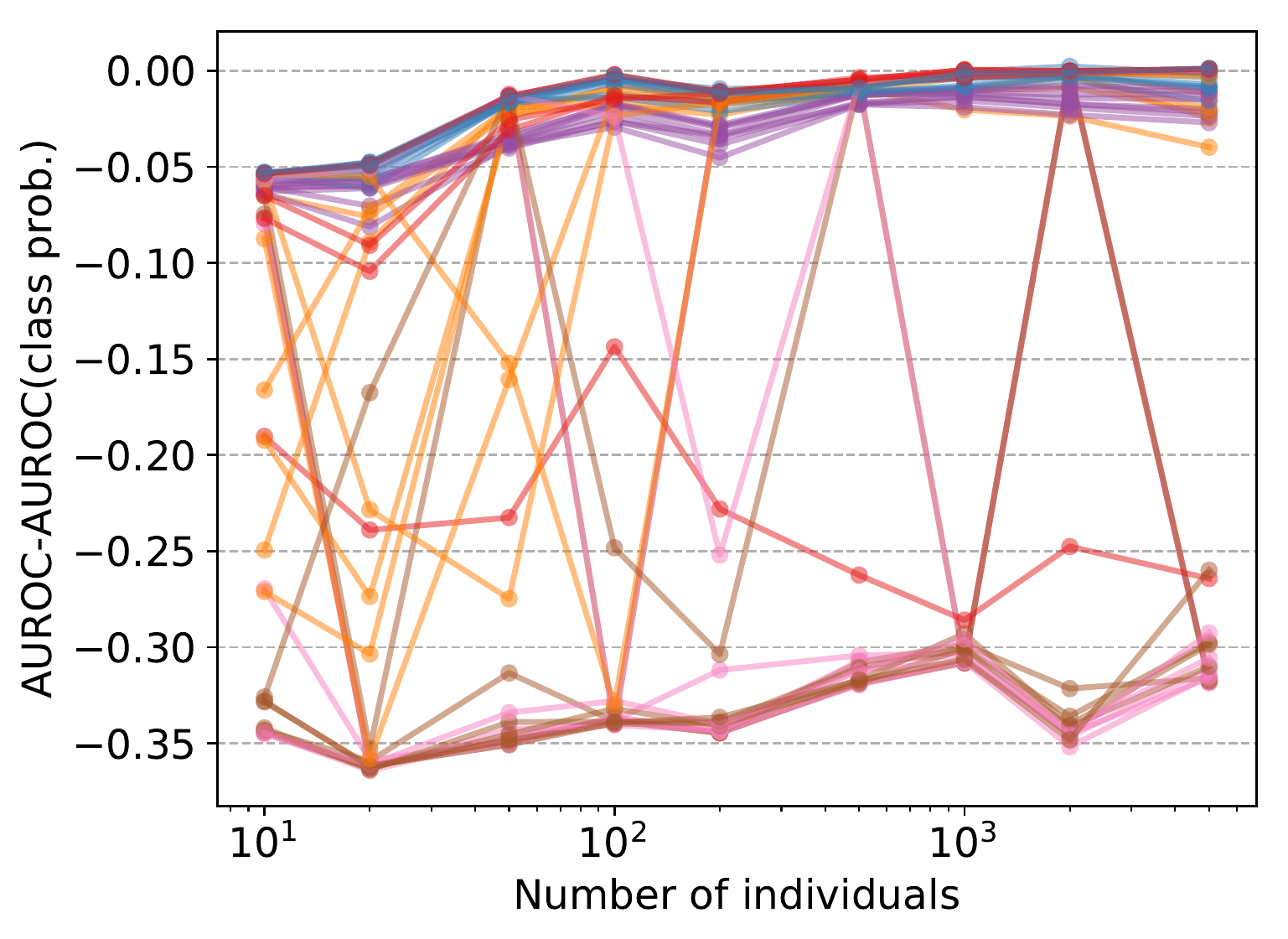}
\end{center}
\caption{\textbf{The AUROC difference between each crowd wisdom and the class probability at different numbers of individuals.} For color legends, see \refig{4}\textbf{D}.\label{fig-converge}}
\end{figure}

\begin{figure}
\begin{center}
See external file: supsimauc.pdf
\end{center}
\caption{\textbf{AUROC and AUPR of dimension reduction methods on different simulated datasets.} \textbf{(A)} to \textbf{(T)}: simulations 2 to 20 and 26 respectively.\label{fig-supsimauc}}
\end{figure}

\begin{figure}
\begin{center}
See external file: supsim1.pdf
\end{center}
\caption{\textbf{Simulation 1 confirmed superior and consistent performances of dimension reduction methods, especially PCA and Isomap.} (\textbf{A}) Comparison of dimension reduction, individual predictions, and the class probability in AUROC and AUPR (cf \refig{2}\textbf{A}) for simulation 1 (\refstab{sim}). (\textbf{B}) Comparison of dimension reduction and supervised learning in cross validation at 25\% training data (cf \refig{3}) for simulation 1. Color reflects relative ranking. \textbf{(C, D, E, F)} ROC (\textbf{C, E}) and Precision-Recall (\textbf{D, F}) curves for dimension reductions, existing crowd wisdoms, the class probability, and individual predictions of simulation 1. In \textbf{C, D}, the best parameter (in \refig{4}\textbf{A}) was selected according to AUROC (\textbf{C}) or AUPR (\textbf{D}) for each parametric dimension reduction and PCA was selected for non-parametric dimension reduction. All methods are shown in \textbf{E, F}. Existing crowd wisdoms were performed on binarized input data.\label{fig-sim1}}
\end{figure}

\begin{figure}
\begin{center}
See external file: supsima5.pdf
\end{center}
\caption{\textbf{Crowd wisdom outperformed supervised learning in cross-validation in simulated dataset.} Empirical distributions and medians of AUROC (left) and AUPR (right) of all crowd wisdom and supervised learning methods in 200 cross-validations with 10\%, 20\%, 25\%, 40\%, 60\%, 80\%, or 90\% (\textbf{A} to \textbf{G}) random partitions of training data are shown for simulation 1.\label{fig-supsima5}}
\end{figure}

\begin{table}
\begin{center}
\caption{\textbf{The AUROC and AUPR of the best individual, neural network, and different crowd wisdom methods.} Numbers in method names represent the number of nearest neighbors.\label{tab-auc}}
\vspace{2em}
See tabc.xlsx.
\end{center}
\end{table}

\begin{table}
\begin{center}
\caption{\textbf{Simulation parameters.}\label{tab-sim}}
\vspace{2em}
See sim.xlsx.
\end{center}
\end{table}

\end{document}